\documentclass{article}

\usepackage{microtype}
\usepackage{graphicx}
\usepackage{float}
\usepackage{caption}
\usepackage{subcaption}
\usepackage{booktabs} % for professional tables
\usepackage{multirow}
\usepackage[accepted]{icml2024}

\usepackage{hyperref}

\usepackage{amsmath}
\usepackage{amssymb}
\usepackage{mathtools}
\usepackage{amsthm}
\usepackage{bbm}
\usepackage[normalem]{ulem}
\usepackage[capitalize,noabbrev]{cleveref}

\theoremstyle{plain}
\newtheorem{theorem}{Theorem}[section]

\newtheorem{corollary}[theorem]{Corollary}
\theoremstyle{definition}
\newtheorem{definition}[theorem]{Definition}

\theoremstyle{remark}
\newtheorem{remark}[theorem]{Remark}
\newcommand{\squishlisttwo}{
 \begin{list}{\scalebox{0.6}{$\bullet$}} 
  { \setlength{\itemsep}{1pt}
     \setlength{\parsep}{0pt}
    \setlength{\topsep}{0pt}
    \setlength{\partopsep}{0pt}
    \setlength{\leftmargin}{1em}
    \setlength{\labelwidth}{1.5em}
    \setlength{\labelsep}{0.5em} } }
    
\newcommand{\squishend}{
  \end{list} 
}

\usepackage[textsize=tiny]{todonotes}

\icmltitlerunning{Helpful or Harmful Data? Fine-tuning-free Shapley Attribution for Explaining Language Model Predictions}

\begin{document}

\twocolumn[

\icmltitle{Helpful or Harmful Data? \\Fine-tuning-free Shapley Attribution for Explaining Language Model Predictions}
\icmlsetsymbol{equal}{*}

\begin{icmlauthorlist}
\icmlauthor{Jingtan Wang}{equal,nus,astar}
\icmlauthor{Xiaoqiang Lin}{equal,nus}
\icmlauthor{Rui Qiao}{equal,nus}
\icmlauthor{Chuan-Sheng Foo}{astar}
\icmlauthor{Bryan Kian Hsiang Low}{nus}
\end{icmlauthorlist}

\icmlaffiliation{nus}{Department of Computer Science, National University of Singapore}
\icmlaffiliation{astar}{Institute for Infocomm Research (I2R), Agency for Science, Technology and Research (A*STAR)}

\icmlcorrespondingauthor{Bryan Kian Hsiang Low}{lowkh@comp.nus.edu.sg}
\icmlkeywords{Machine Learning, ICML}

\vskip 0.3in
]

\printAffiliationsAndNotice{\icmlEqualContribution} % otherwise use the standard text.

\begin{abstract}

The increasing complexity of foundational models underscores the necessity for explainability, particularly for fine-tuning, the most widely used training method for adapting models to downstream tasks. Instance attribution, one type of explanation, attributes the model prediction to each training example by an instance score. However, the robustness of instance scores, specifically towards dataset resampling, has been overlooked. To bridge this gap, we propose a notion of robustness on the sign of the instance score. We theoretically and empirically demonstrate that the popular leave-one-out-based methods lack robustness, while the Shapley value behaves significantly better, but at a higher computational cost. Accordingly, we introduce an efficient \underline{f}ine-tuning-f\underline{ree} approximation of the \underline{Shap}ley value (FreeShap) for instance attribution based on the neural tangent kernel. We empirically demonstrate that FreeShap outperforms other methods for instance attribution and other data-centric applications such as data removal, data selection, and wrong label detection, and further generalize our scale to large language models (LLMs). Our code is available at \url{https://github.com/JTWang2000/FreeShap}.

\end{abstract}

\section{Introduction}
Modern deep learning is primarily driven by pretraining foundation models \citep{bommasani2021opportunities} such as BERT~\citep{devlin-etal-2019-bert}, GPT3~\citep{brown2020language}, and CLIP~\citep{radford2021learning} on massive datasets with increasingly more model parameters, then fine-tuning them with labeled training data for downstream tasks. However, as larger foundation models have more complexity and opaqueness, their predictions are hard to explain and justify. The lack of explainability reduces the trustworthiness of foundation models, especially for systems in areas that involve high-stakes decision-making such as healthcare, finance, and legal justice \citep{Rudin2019}.
As a result, \textit{instance attribution}, also known as example-based explanation, has been proposed in response to the increasing need for explainable AI. Different from other types of explanation (e.g., feature attribution \citep{li-etal-2016-visualizing, madsen2022post, ribeiro-etal-2016-trust, sundararajan2017axiomatic}), instance attribution traces the effect of the labeled training examples on the resultant model parameters~\citep{pmlr-v108-barshan20a, guo-etal-2021-fastif, han-etal-2020-explaining, pezeshkpour-etal-2021-empirical}. 
On a high level, this approach offers a better understanding of the fine-tuned model by sifting the influential training examples with larger effects on the model performance. 
On a detailed level, it explains the model prediction on each data point by highlighting the most relevant training examples. Besides, instance attribution can also be practically used for data selection~\citep{xia2024less}, the identification of dataset artifacts~\citep{han-tsvetkov-2021-influence-tuning}, model debugging~\citep{lertvittayakumjorn-toni-2021-explanation}, wrong label detection, adversarial example crafting \citep{koh2017understanding}, and data pricing in collaborative machine learning~\citep{ijcai2022p782}.

\begin{figure}[h]
\begin{center}
\centerline{\includegraphics[width=\columnwidth]{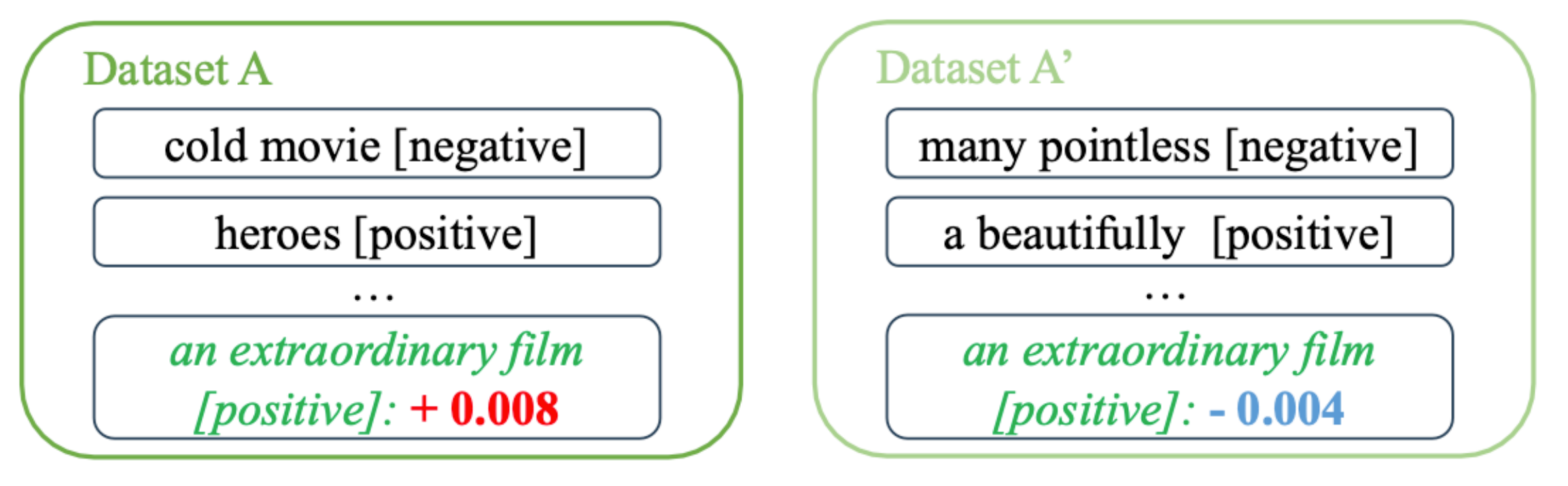}}
\caption{An example of non-robust instance attribution. The \textit{same} training example receives \textit{different} signs of the instance score when it is placed in different datasets sampled from the same task.}
\label{fig: nonrobust}
\end{center}
\end{figure}

Given a dataset for a downstream task, instance attribution calculates an instance score for each training example. A positive instance score deems the example ``helpful'' for the task as it improves the model performance, while a negative score reflects the opposite. Thus, the sign of the scores is a widely used criterion for the usefulness of the data~\citep{han-etal-2020-explaining,yeh2018representer}. 
It is important to realize that approaches for instance attribution are dataset-dependent, meaning that the instance scores can be different when the dataset is resampled. If the instance score for one example flips signs frequently when the rest of the training examples change as shown in Fig.~\ref{fig: nonrobust},  this score results in a confusing and less trustworthy analysis. 
Ideally, the instance score for a ``helpful'' example should be consistently positive even if the rest of the training data are resampled from the same task, the same applies to a ``harmful'' example.
Thus, we define the \textit{robustness of instance attribution} as the ability of an approach to maintain the sign consistency of the instance scores across different resampled datasets. 
Our robustness definition is motivated from the perspective of data distribution, hence complementary to existing perturbation-based robustness measures~\citep{alvarez2018robustness, ijcai2020p417, Ghorbani_Abid_Zou_2019, tang-etal-2022-identifying}.

\begin{figure}[h]
\begin{center}
\centerline{\includegraphics[width=\columnwidth]{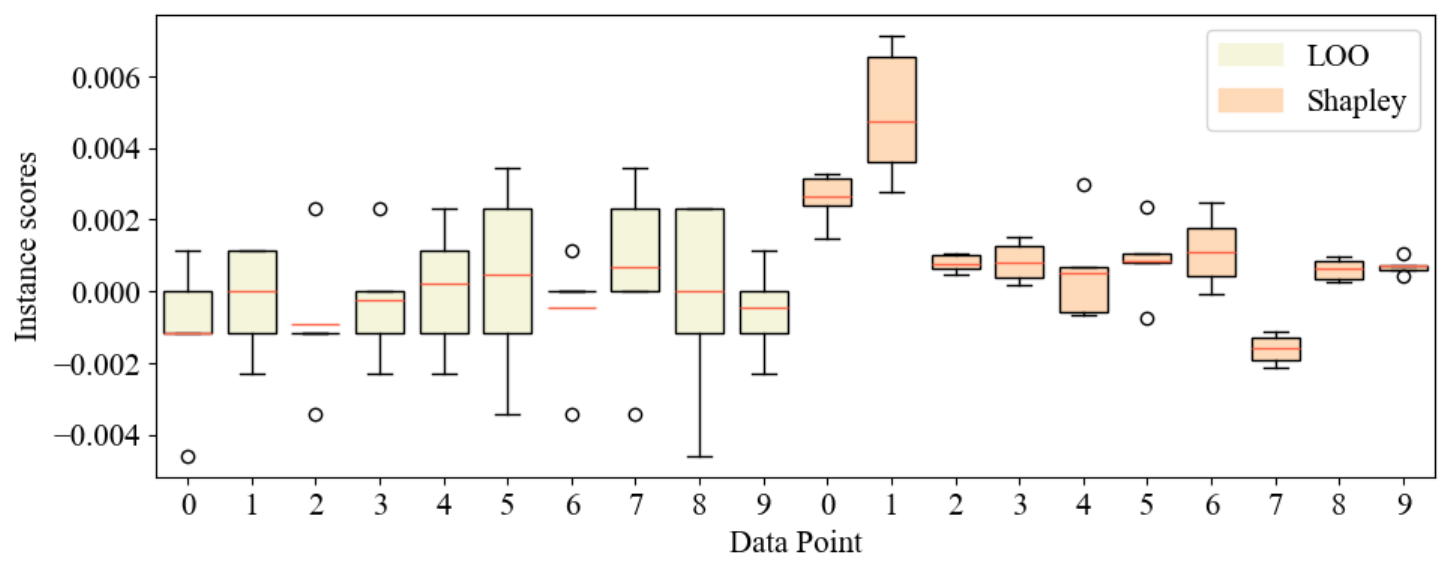}}
\caption{Mean and variance for instance scores of 10 examples when computed using LOO or the Shapley value.}
\label{fig: intuition}
\end{center}
\end{figure}

Most of the popular instance attribution approaches~\citep{pmlr-v108-barshan20a, han-etal-2020-explaining, guo-etal-2021-fastif, pruthi2020estimating} are based on the leave-one-out (LOO) scheme, which evaluates the marginal contribution (i.e., the gain in model performance by adding the target data point) to the rest of the training set. 
However, LOO tends to produce instance scores with small magnitude and high variance~\citep{pmlr-v151-kwon22a}. 
Therefore, LOO-based scores are more likely to have inconsistent signs when the dataset is resampled (see Tab.~\ref{table: sst2_robust}), leading to poor robustness. On the contrary, the Shapley value~\citep{pmlr-v97-ghorbani19c, pmlr-v89-jia19a, pmlr-v206-wang23e} considers a weighted average of marginal contributions to all training subsets. As a result, it tends to enjoy a relatively larger magnitude and smaller variance (see Fig.~\ref{fig: intuition}). Moreover, we theoretically and empirically show that, under mild assumptions, the Shapley value possesses noticeably better robustness compared to LOO in Sec.~\ref{sec: robust_instance_attribution} and Sec.~\ref{sec: robust_experiment}.

However, the time complexity to compute the exact Shapley value grows exponentially w.r.t. the number of training data, which is substantially amplified by the computational cost of fine-tuning. 
Although existing approaches have attempted to reduce the exponential time w.r.t. dataset size~\citep{pmlr-v97-ghorbani19c, pmlr-v89-jia19a, kolpaczki2023approximating}, 
the cost of fine-tuning has been overlooked.
As the size of foundation models grows larger, there is an increasing need to reduce the running time for fine-tuning in order to efficiently approximate the Shapley value. To this end, we propose an efficient \underline{f}ine-tuning-f\underline{ree} approximation of the \underline{Shap}ley value (FreeShap).
Specifically, it has been shown that kernel regression using empirical neural tangent kernel (eNTK) ~\cite{NEURIPS2018_5a4be1fa} resembles the fine-tuning process~\citep{pmlr-v162-wei22a, pmlr-v202-malladi23a}. 
An observation is that the eNTK matrices for different data subsets are submatrices of the eNTK matrices of the whole dataset. We exploit this observation to accelerate FreeShap. Specifically, FreeShap amortizes the cost of calculating the Shapley values by pre-computing the eNTK matrix to replace all subsequent training with kernel regression (i.e., fine-tuning-free) which is significantly cheaper in computational cost than fine-tuning. 
Our approach generally applies to \textit{any} foundation model, but we focus our demonstrations in the natural language processing (NLP) domain in light of the breakthroughs in language models (LMs), especially LLMs.
Our contributions are:

\squishlisttwo
\item Introducing $\beta$-robustness, a notion of robustness for \textit{instance attribution} (Sec.~\ref{sec: def_robust}), and theoretically analyzing and empirically demonstrating that the Shapley value is more robust than LOO (Sec.~\ref{sec: robust_instance_attribution} and Sec.~\ref{sec: robust_experiment}).
\item Proposing an efficient fine-tuning-free approximation of Shapley value, FreeShap, by \textit{replacing fine-tuning with kernel regression}, increasing the scalability of the Shapley value significantly for NLP tasks (Sec.~\ref{sec: entk-shapley} and Sec.~\ref{sec: corr_exp}).
\item Demonstrating the effectiveness of the FreeShap through extensive experiments using pre-trained LMs and LLMs for instance attribution, data removal, data selection, and wrong label detection on real-world datasets (Sec.~\ref{sec: application_instance_attribution}).
\squishend

\section{Background and Preliminaries}
Let $D_N \coloneq \{z_i = (x_i, y_i)\}_{i=1}^n$ denote the training set, where $N \coloneq \{1, \dots, n\}$ is the set of indices, and $z_i$ consisting of the input $x_i \in \mathcal{X}$ and the label $y_i \in [C]$ is a data point sampled from a data distribution $\mathcal{P}$. Similar notations apply to the test set $D_{T} \coloneq \{z_t = (x_t, y_t)\}_{t=n+1}^{n+m}$, the test indices $\textstyle T \coloneq \{n+1, \dots, n+m\}$, and the test example $z_t$.
An attribution function $g(z_i, D_T, D_N)$ quantifies the contribution of a training example $z_i$ to the model predictions on the test set $D_T$ (or the test example $\{z_t\}$). 
We define the utility for a subset of the training data $D_S:= \bigcup_{j \in S} \{z_j \in D_N\}, S \subseteq N$ as the test accuracy on $D_T$ of its resultant model $f_S$ parameterized by $\theta$: 
\begin{equation}
\label{eq: utility}
    U(S, D_T) \coloneq \textstyle \frac{1}{|D_T|} \sum_{(x_t,y_t)\in D_T} \mathbbm{1}[f_S(x_t; \theta)=y_t]\ .
\end{equation}

\subsection{Prompt-based Fine-tuning}
\label{preliminary: prompted_finetuning}
The LMs are pretrained on tasks such as masked token prediction~\citep{devlin-etal-2019-bert} or next-token prediction~\citep{brown2020language}.
Prompt-based fine-tuning~\citep{gao-etal-2021-making, schick-etal-2021-self} is an advanced technique that appends natural language cues like ``This is [MASK].'' (i.e., prompt) to the end of $x$ and then employs the pretrained model to predict the word for [MASK] from words that are relevant to the task. For example, binary sentiment analysis can be formulated as predicting only the task-specific terms including ``great'' and ``terrible'' at the ``[MASK]'' location. Prompt-based fine-tuning has been shown to have better performance than fine-tuning with sentence representations (e.g., the last hidden-layer representation for the [CLS] token) without prompts. Consequently, we focus on prompt-based fine-tuning in this work. 

\subsection{LOO and Shapley Value for Instance Attribution}
Most of the instance attribution approaches can be viewed as some forms of marginal contributions \citep{pmlr-v151-kwon22a}. Define the marginal contribution of a data point $z_i$ to the subset with size $k$ of all other training data as:
\begin{equation}\label{eq: marginal_contribute}
\resizebox{1.0\linewidth}{!}{$ 
    \Delta_{z_i}^{D_N}(k,D_T) \coloneq \frac{1}{\binom{n-1}{k}} \sum_{\substack{S \subseteq N \setminus \{i\} \\ |S|=k}} U(S \cup \{i\}, D_T) - U(S, D_T)
$} 
\end{equation}
Since LOO \citep{koh2017understanding} considers the change in model accuracy by removing the target data point from the training set, it can be viewed as the marginal contribution of $z_i$ to $D_N\backslash \{z_i\}$. Specifically, LOO can be written as the following instance scoring function:
\begin{equation}
\begin{aligned}
\label{eq: LOO}
    g^{\text{LOO}}(z_i, D_T, D_N) &= \Delta_{z_i}^{D_N}(n-1,D_T)\ . 
\end{aligned}
\end{equation}
Many existing instance attribution schemes including the influence function~\citep{han-etal-2020-explaining} and its subsequent works~\citep{pmlr-v108-barshan20a, guo-etal-2021-fastif} approximate the LOO scores in different ways. However, LOO only considers the marginal contribution of $z_i$ to one subset of $D_N$ with size $n-1$. 
The instance score of an example drops significantly when another similar example appears in the training set \citep{pmlr-v97-ghorbani19c}, leading to the weak reliability and robustness of the score.

Instead of considering only one subset, the Shapley value~\citep{shapley1953value} generalizes it by considering the impact of $z_i$ in all subsets~\citep{pmlr-v97-ghorbani19c} and can be written as the following instance scoring function:
\begin{equation}
\label{eq: shap}
\begin{aligned}
    g^{\text{Shap}}(z_i, D_T, D_N) &= \textstyle n^{-1} \sum_{k=0}^{n-1} \Delta_{z_i}^{D_N}(k,D_T) \ .
\end{aligned}
\end{equation}
This can be viewed as the average of the marginal contributions of data point $z_i$ to subsets with all possible sizes. However, calculating the Shapley value on the training set with size $n$ requires an exponential number (i.e., $n!$) of LM fine-tuning, which necessitates acceleration for improved scalability on real-world datasets and models.

\subsection{Neural Tangent Kernel}
\label{sec: ntk}
The neural tangent kernel (NTK) theory is proposed to study the training dynamics of infinite-width neural networks ~\citep{NEURIPS2018_5a4be1fa}
, and has been extended to other applications, such as active learning~\citep{hemachandra2023training, lau2024pinnacle}. 
Existing works~\citep{NEURIPS2019_dbc4d84b,liu2020toward} show that training a fully connected and sufficiently wide neural network is equivalent to solving kernel regression with the NTK at random initialization. 
However, there are two challenges when applying the NTK theory to fine-tuning LMs:
(1) the prevalence of using pretrained weights instead of random initialization; (2) the usage of prompts. 
\citet{pmlr-v202-malladi23a} extends the analysis to show that solving the kernel regression with the empirical NTK (eNTK) calculated from the pretrained weights can resemble prompt-based fine-tuning. Moreover, kernel regression using eNTK has been shown to have similar performance as fine-tuning in both computer vision~\citep{pmlr-v162-wei22a} and NLP~\citep{pmlr-v202-malladi23a}, which indicates its applicability as a surrogate to fine-tuning.

Specifically, the eNTK is calculated using the Jacobian of the model output when taking a data point $x_i$: $\psi(x_i) \coloneqq \frac{\partial f (x_i; \theta_0)}{\partial \theta_0} \in \mathbb{R}^{C \times P}$, where $\theta_0 \in \mathbb{R}^P$ consists of the pretrained weights.
For a training dataset $D_S$ with indices $S \coloneqq \{1, \dots, k\}$, denote the input matrix by $X_S \coloneqq [x_1, \dots, x_k]^\top$ and the corresponding labels in one-hot by $ Y_S \coloneqq [y_1^1, \dots, y_1^C, \dots, y_k^1, \dots, y_k^C]^\top  \in \{0,1\}^{kC}$, where $y_k^j$ is $1$ if $y_i=j$. A test input $x_t$ can be predicted using the eNTK regression model as follows:
\begin{equation}
\label{eq:kernel_regression}
    f^{\text{entk}}_S(x_t) = K(x_t, X_S)^\top K(X_S, X_S)^{-1}Y_S \
\end{equation}
where ${K(x_t, X_S)} = {\bigl[ \psi(x_t) \psi(x_i)^\top \bigr]}_{i=1}^k \in \mathbb{R}^{kC \times C}$ and $K(X_S, X_S) = {\bigl[ \psi(x_i) \psi(x_j)^\top \bigr]}_{i, j=1}^k \in \mathbb{R}^{kC \times kC}$.

\section{Methodology}
\subsection{Definition of Robustness}
\label{sec: def_robust}
Denote $D_N \sim \mathcal{P}^{n-1}|z_i$ as sampling a new dataset $D_N$ with $z_i$ fixed and other $n-1$ data points i.i.d. sampled from $\mathcal{P}$.
\begin{definition}[Expected marginal contribution] Let $\tau_k \coloneqq \mathbb{E}_{D_N \sim \mathcal{P}^{n-1}|z_i}\bigl[\Delta_{z_i}^{D_N}(k,D_T)\bigl]$ be the expected marginal contribution of $z_i$ to the subsets of $D_N$ with size $k$.
\end{definition}

\begin{definition}[Consistently helpful/harmful data point]
A training data point $z_i$ is \textit{consistently helpful} to test set $D_T$ if
$\tau_k \geq 0, \forall k \geq \{0, \dots, n-1\}$. Similarly, it is \textit{consistently harmful} if $\tau_k < 0$.
\end{definition}
If $z_i$ is a \textit{consistently harmful} data point, including $z_i$ in any subset of the dataset is expected to hurt the model performance on the test set.
Intuitively, a \textit{consistently harmful} data point $z_i$ should receive a negative instance score. However, due to the randomness of the other data points when sampling the dataset, the attribution score might flip to non-negative. 
This is undesirable, especially for high-stakes applications, since a harmful data point can be interpreted as helpful due to its positive score at one specific dataset and hence not being removed. 
Therefore, it is preferable to have a robust instance attribution scheme that could assign the correct signs for the training examples more consistently regardless of the companion dataset $D_N$.
We emphasize the \textit{sign} as it is the natural threshold that determines the usefulness of the data~\citep{han-etal-2020-explaining, yeh2018representer}.
Therefore, we define the following:
\begin{definition}[Robustness of instance attribution]
\label{def: robust}
A sign function is defined as \( \text{sgn}(x) = -1 \) if \( x < 0 \), and \( \text{sgn}(x) = 1 \) if \( x \ge 0 \). For a \textit{consistently harmful} point $z_i$, let the $\text{sgn*}(z_i)=-1$ , while for a \textit{consistently helpful} point $z_i$, let the $\text{sgn*}(z_i)=1$. An instance attribution approach is $\beta$-robust in giving instance score for a data point if:
\begin{equation*}
\mathbb{P}_{D_N \sim \mathcal{P}^{n-1}|z_i}\biggl( \text{sgn}\Bigl(g(z_i, D_T, D_N)\Bigl) \neq \text{sgn*}(z_i) \Bigl) \biggl) = \textstyle \beta.
\end{equation*}
\end{definition}
In other words, given a training example $z_i$, an instance attribution scheme is $\beta$-robust if it has a probability of $\beta$ in giving an instance score with the opposite sign when resampling the dataset. 
Overall, a robust instance attribution scheme should maintain a small $\beta$ and ensure the \textit{sign}-consistency of the instance scores for a data point across different datasets drawn from the same data distribution.

\subsection{Robustness of Instance Attribution}
\label{sec: robust_instance_attribution}
\begin{theorem}[Robustness for Shapley value \& LOO]
\label{theorem:robust-sv-loo}
Let \\$\delta_k \coloneqq \text{Var}_{D_N \sim \mathcal{P}^{n-1}|z_i}(\Delta_{z_i}^{D_N}(k,D_T)) , \forall k \in \{0, \dots, n-1\}$.
Shapley value is $\beta^{\text{Shap}}$-robust and LOO is $\beta^{\text{LOO}}$-robust where 
$$\beta^{\text{Shap}} \le  \frac{n^{-1} \sum_{k=0}^{n-1} \delta_k}{(n^{-1} \sum_{k=0}^{n-1} \tau_k)^2} \quad and \quad \beta^{\text{LOO}} \le  \frac{\delta_{n-1}}{\tau_{n-1}^2}\ .$$
\end{theorem}
The proof is in App.\textcolor{red}{~\ref{appendix:proofs}}.
On the variance $\delta_k$, 
Theorem 1~\citep{pmlr-v151-kwon22a} shows that the asymptotic of $\delta_k \sim O(Ck^2/n)$ (i.e., the upper bound of the $\delta_k$ scales quadratically with the subset size, $k$), it is reasonable to assume that $n^{-1}\sum_{k=0}^{n-1}\delta_k \le \delta_{n-1}$. 
Additionally, as the dataset size grows, the contribution of each training example on the model typically diminishes, because the abundance of other data points is more likely to compensate for the performance drop. 
Hence, we also assume diminishing influence, i.e., $|\tau_0| \geq \dots\geq |\tau_{n-1}|$, which coincides with the assumptions in~\citet{killamsetty2021glister,wang2021unified} and is empirically verified in App.~\ref{app: diminishing_mc}. As a result:
\begin{corollary}[Robustness Analysis between the Shapley value and LOO]
Assume that $n^{-1}\sum_{k=0}^{n-1}\delta_k \le \delta_{n-1}$. Additionally, for any \textit{consistently harmful (or helpful) contributing} data point $z_i$, assume $|\tau_0| \geq \dots\geq |\tau_{n-1}|$.
Then, 
\begin{equation}
    \frac{n^{-1} \sum_{k=0}^{n-1} \delta_k}{(n^{-1} \sum_{k=0}^{n-1} \tau_k)^2} \leq  \frac{\delta_{n-1}}{\tau_{n-1}^2}.
\end{equation}
\end{corollary}
This implies that the upper bound of the $\beta$-robustness of the Shapley value is no more than that of LOO. The upper bounds are valid proxies to the $\beta$-robustness for comparison because they are \textit{derived with the same non-trivial techniques}. Having a better upper bound suggests that the Shapley value can be less likely to give sign-inconsistent scores, and hence more robust. Note that the robustness analysis applies to \emph{every} training example $z_i$.
We provide empirical evidence to demonstrate that the Shapley value is indeed more robust than LOO in Sec.~\ref{sec: robust_experiment}, w, which further validates our theoretical insights.

The numerator in the upper bounds of $\beta$ is related to the variance of the instance score, while the denominator is related to the expectation of the instance score. Thus, the robustness of the Shapley value can also be interpreted from the expectation and variance. 
Without loss of generality, we analyze for a \textit{consistently helpful} data point.
\begin{remark}[Relative relationship of expectation and variance between Shapley and LOO]
\label{eqn: remark}
For a \textit{consistently helpful} point $z_i$, with the assumption that $|\tau_0| \geq \dots\geq |\tau_{n-1}|$: 
\begin{equation}
    \begin{aligned}
    \mathbb{E}_{D_N \sim \mathcal{P}^{n-1}|z_i}\bigl(g^{\text{Shap}}(z_i, D_T, D_N)\bigl) = \textstyle n^{-1}\sum_{k=0}^{n-1} \tau_k \\
    \geq \tau_{n-1} = \mathbb{E}_{D_N \sim \mathcal{P}^{n-1}|z_i}\bigl(g^{\text{LOO}}(z_i, D_T, D_N)\bigl)\ .
\end{aligned}
\end{equation}
With the assumption that $n^{-1} \sum_{k=0}^{n-1}\delta_k \le  \delta_{n-1}$: 
\begin{equation}
\begin{aligned}
     \text{Var}_{D_N \sim \mathcal{P}^{n-1}|z_i}\bigl(g^{\text{Shap}}(z_i, D_T, D_N)\bigl) \leq \textstyle n^{-1}\sum_{k=0}^{n-1}\delta_k \\
     \le  \delta_{n-1}= \text{Var}_{D_N \sim \mathcal{P}^{n-1}|z_i}\bigl(g^{\text{LOO}}(z_i, D_T, D_N)\bigl)\ .
\end{aligned}
\end{equation}
\end{remark}
Overall, the Shapley value can have a larger magnitude of expectation and smaller variance than LOO. For a \textit{consistently helpful} data point, the Shapley value has a lower probability of fluctuating around the borderline $0$ due to the larger expectation and smaller variance. The remark is further empirically verified in the App.~\ref{app: mean_std_robust}.

\subsection{Fine-tuning-free Shapley Value (FreeShap)}
\label{sec: entk-shapley}
Shapley value computation requires marginal contribution computations.
For each subset $S$, we need to retrain a model $f_S$. As there are $2^n$ subsets for $D_N$, exponential times of fine-tuning are needed. Prior works have adopted Monte-Carlo (MC) sampling~\citep{10.1016/j.cor.2008.04.004, maleki2013bounding, pmlr-v97-ghorbani19c, lin2023fair} to avoid the exponential factor. 
Specifically, the marginal contribution of each data point is evaluated in each MC iteration where a random permutation is generated. The average result of each iteration is the final estimation of the Shapley value. MC sampling gives an unbiased estimate of the Shapley value with a tolerable error bound~\citep{maleki2013bounding}. 

Although MC sampling has reduced the exponential time to polynomial time, it is still a large factor that multiplies with the fine-tuning cost. Fine-tuning a large pre-trained LM itself is computationally expensive. To further speed up the Shapley value for large NLP datasets, we aim to improve the complexity of computing the utility function which relies on the fine-tuned model $f_S$.
Instead of fine-tuning, we adopt kernel regression on the \textit{empirical} NTK (eNTK)~\citep{pmlr-v162-wei22a}, using $f_S^{\text{entk}}$ to resemble $f_S$. Note that $D_N$ denotes the training dataset of size $n$, while $D_T$ denotes the testing dataset of size $m$, and $D_S$ denotes a subset of the training dataset.
We just need to precompute the eNTK matrix $\tilde{K}$ once, as shown in line 4 of Alg.~\ref{algo: training-free}, and then reuse it for all marginal contributions by taking the submatrices of $\tilde{K}$ for corresponding kernel regression. Specifically, 
\begin{equation}\label{equ:slice-operation}
\resizebox{0.91\linewidth}{!}{$
    \tilde{K}{[T,S^t_p]}=K(X_{T}, X_{S^t_p}), \quad \tilde{K}[S^t_p, S^t_p]=K(X_{S^t_p}, X_{S^t_p}) .$
}
\end{equation}
We refer to this method as the fine-tuning-free Shapley value (FreeShap), which ensures the scalability of computing the Shapley value with pretrained LMs on sizable datasets. We further enhance the computational efficiency by truncated Monte-Carlo (TMC) sampling~\cite{Ghorbani_Abid_Zou_2019} and employing efficient kernel approximation and blockwise inversion (BI), which are discussed in App.~\ref{app: time_TFSV}. 
\begin{algorithm}[tb]
    \caption{Fine-tuning-free Shapley Value (FreeShap)}
\begin{algorithmic}[1]
\label{algo: training-free}
\STATE {\bfseries Require:} Training set $D_N$, test set $D_T$, utility function $U$, maximum MC iteration $M$, eNTK function $K$\;
\STATE {\bfseries Output:} Instance score of training examples computed by the Shapley value: $\phi_1, \dots, \phi_n$\;
\STATE Initialize $\phi_i \leftarrow 0$ for $i = 1, \dots, n$, $t \leftarrow 0$\;
\STATE Precompute the eNTK matrix $ \tilde{K} \coloneq K(X_{T \cup N}, X_N)$
\WHILE{$t \leq M$}
    \STATE $t \leftarrow t + 1$\;
    \STATE $\pi^t$: A permutation of the training indices $N$\;
    \STATE $u^t_0 \leftarrow U(\emptyset, D_T)$\;
    \FOR{$p \in \{1, \dots, n\}$}
        \STATE $S^t_p \leftarrow \{\pi^t[1], \dots, \pi^t[p]\}$\;
         \STATE $f_{S^t_p}^{\text{entk}}(D_T) = \tilde{K}{[T,S^t_p]} \tilde{K}[S^t_p, S^t_p]^{-1} Y_{S^t_p}$\\
         \COMMENT{See Eqn.~\eqref{equ:slice-operation}} 
        \STATE $u^t_p \leftarrow U(S^t_p, D_T|f_{S^t_p} = f_{S^t_p}^{\text{entk}})$
        \STATE $\phi_{\pi^t[p]} \leftarrow \frac{t-1}{t} \phi_{\pi^{t-1}[p]} + \frac{1}{t} (u^t_p - u^t_{p-1})$\;
    \ENDFOR
\ENDWHILE
\end{algorithmic}
\end{algorithm}

\section{Experiments and Results}
\label{sec: experiment}
We first empirically demonstrate that the proposed FreeShap effectively approximates the de facto implementation of the Shapley value. 
Hence, it justifies the usage of FreeShap for all subsequent experiments at scale. We then empirically compare the $\beta$-robustness between the FreeShap and LOO and show that the Shapley value is more robust, which is aligned with our theoretical insights in Sec.~\ref{sec: robust_instance_attribution}.

\textbf{Datasets}: We conduct experiments on both single-sentence tasks and sentence pair tasks for comprehensiveness. In particular, we select Stanford Sentiment Treebank v2 (SST-2) and Rotten Tomatoes Movie Review (MR) for the single-sentence task, and Microsoft Research Paraphrase Corpus (MRPC) and Recognizing Textual Entailment (RTE) for the sentence pair task.
Detailed information about the datasets is in App.~\ref{app: dataset}. We use BERT for this experiment. More hyperparameter settings are in the App.~\ref{app: robust_hyperparameter}. 

\subsection{FreeShap Approximates the Shapley Value Well}
\label{sec: corr_exp}

\textbf{Setup.}
Due to the exponential time complexity for the exact Shapley value, we use the Monte-Carlo (MC) Shapley value as the reference. We perform prompt-based fine-tuning with the Adam optimizer~\citep{kingma2014adam} when calculating its marginal contribution. We benchmark against gradient Shapley (G-Shapley)~\citep{pmlr-v97-ghorbani19c}, which also minimizes fine-tuning costs by limiting training to a single epoch. 
For evaluation metrics, we utilize Pearson correlation coefficients to capture linear relationships and Spearman correlation coefficients to measure the degree of similarity between rankings. For each dataset, we randomly sample a subset of 500 training examples, then measure the similarity of the Shapley values calculated by different methods on 50 random data points from the subset. The scale is limited here because the reference algorithm MC-Shapley is extremely slow. 
All experiments are repeated for 3 trials.

\begin{table}[ht]
\caption{Correlations with MC-Shapley when the dataset size is 500. The higher, the better.}
\resizebox{1.0\linewidth}{!}{
\centering
\begin{tabular}{c|cc|cc}
\toprule
\multirow{2}{*}{Dataset} & \multicolumn{2}{c|}{FreeShap} & \multicolumn{2}{c}{G-shapley}\\
 &  Pearson & Spearman & Pearson & Spearman \\
\midrule
\multicolumn{5}{c}{Single Sentence Task}\\
\midrule
SST-2  & \textbf{0.70$\pm$0.08}  &  \textbf{0.60$\pm$0.03}  &  0.44$\pm$0.08  &  0.48$\pm$0.14 \\ 
MR    & \textbf{0.57$\pm$0.15}  &  \textbf{0.53$\pm$0.04}  &  0.32$\pm$0.22  &  0.32$\pm$0.23 \\
\midrule
\multicolumn{5}{c}{Sentence Pair Task}\\
\midrule
MRPC  & \textbf{0.84$\pm$0.04}  &  \textbf{0.73$\pm$0.09}  &  0.68$\pm$0.08  &  0.58$\pm$0.11 \\ 
RTE   & \textbf{0.78$\pm$0.04}  &  \textbf{0.72$\pm$0.02}  &  0.28$\pm$0.04  &  0.36$\pm$0.05 \\ 
\bottomrule 
\end{tabular}
}
\label{table:shapley-corr}
\end{table}

\begin{figure}[htb]
\centering
\includegraphics[width=8cm]{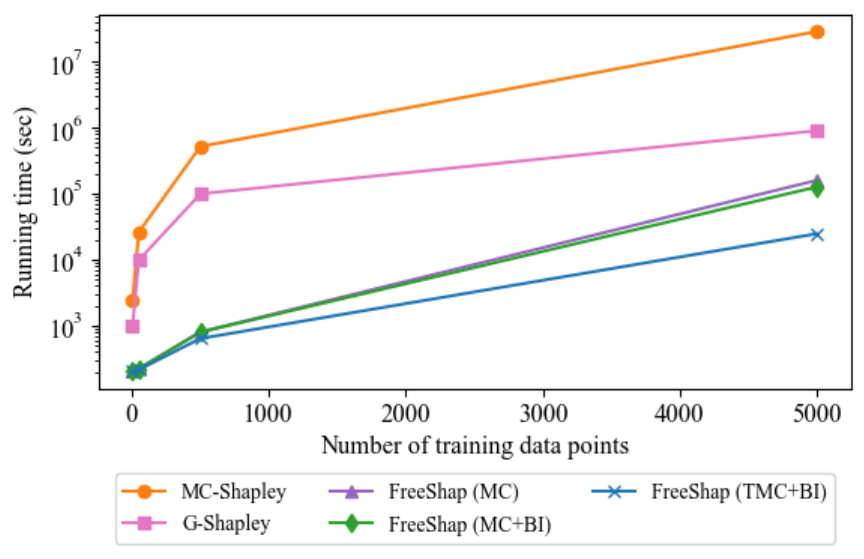}
\caption{Running time comparison. 
The time for 5k points for G-Shapley and 500/5k points for MC-Shapley are projected.
}
\label{figure: time}
\end{figure}

\textbf{Results.}
Tab.~\ref{table:shapley-corr} demonstrates that our approach is consistently more effective in approximating the MC-Shapley compared to G-Shapley on all four NLP datasets. In particular, FreeShap has relatively high positive linear correlations and ranking correlations, which are also positively correlated with each other. 
The competitive correlations of FreeShap persist when the subset size grows to 1000 (App.~\ref{app: more_result_corr}). 

FreeShap is also more computationally efficient than other methods, as shown in Fig.~\ref{figure: time} which estimates the running time on the SST-2 dataset. 
Due to the better efficiency and more reliable performance, we use FreeShap for computing the Shapley value in the following sections.

\subsection{The Shapley Value is More Robust than LOO}
\label{sec: robust_experiment}
\looseness=-1
\textbf{Setup.}
To empirically compare the $\beta$-robustness of the Shapley value and the LOO, we evaluate the consistency of the signs for a data point's instance score when the point lies in different datasets drawn from the same distribution.
Due to the necessity of sampling numerous datasets when determining if a data point is \textit{consistently harmful} or \textit{helpful}, it is impossible to calculate the exact $\beta$. 
To empirically estimate the $\beta$-robustness, we design the experiments as follows: 
For a data point, we place it into 5 sampled training sets. This process yields 5 corresponding instance scores for the data point. We use the majority sign of these scores to determine whether the data point is \textit{consistently harmful} or \textit{helpful}. If any score's sign contradicts the majority, this is noted as a non-robust occurrence. We evaluate 50 training examples with the same 5 complementary dataset samples. We use the percentage of non-robust occurrence out of the 50 training examples as the empirical metric for $\beta$-robustness. An illustrative figure of the setup is in Fig.~\ref{fig: exp_setup} in the Appendix. We approximate the Shapley value with FreeShap and calculate the exact LOO score. The experiment is performed comprehensively on datasets of various sizes.

\textbf{Results.}
Tab.~\ref{table: sst2_robust} shows that for each size of the training set, the Shapley value generally has fewer non-robust occurrences, hence its instance scores are generally more robust than the one from LOO. Results on more dataset sizes are in App.~\ref{app: robust_result}.
We subsequently analyze a few exemplary examples where the Shapley value succeeds but LOO fails in computing consistent instance scores in Tab.~\ref{table: robust_case_study}.
The training example ``it’s not going to be everyone’s bag of popcorn, but it gives you something to chew on'' reflects a positive sentiment, but it lacks clear polarity. 
Including such challenging training points can adversely impact test performance. It might be better to categorize the example as low-quality for further curation.
Nonetheless, LOO-based instance attribution might sometimes identify this sample as helpful. Consequently, users might incorrectly conclude that this sample is beneficial for the task and retain it in the dataset, which impairs the performance. Furthermore, the explanations could influence future data collection, resulting in the accumulation of training instances with ambiguous polarity. 
Overall, the Shapley value yields instance scores that are much more sign-consistent and better aligned with human intuition: negative scores indicate ambiguity, while non-negative scores are associated with better clarity and polarity. These results are further demonstrated in App~\ref{app: robust_result}, where we show more case studies of different dataset sizes.

\begin{table}[h]
\caption{
The empirical comparison of the $\beta$-robustness between the Shapley value and LOO by the percentage of non-robust occurrences. The lower, the better. 
}
\centering
\resizebox{1.0\linewidth}{!}{
\begin{tabular}{c|cc|cc|cc}
\toprule
\multirow{2}{*}{} &\multicolumn{2}{c|}{$|D_N|=500$} &  \multicolumn{2}{c|}{$|D_N|=1000$} & \multicolumn{2}{c}{$|D_N|=2000$} \\
dataset & LOO&Shapley  &  LOO&Shapley  &  LOO&Shapley \\
\midrule
SST-2  & 86\% &\textbf{30\%}  &  \textbf{38\%} & 50\%    & 90\% & \textbf{46\%} \\ 
MR    & 98\% &\textbf{18\%}  &  100\% & \textbf{28\%}   & 100\% & \textbf{44\%} \\
MRPC  & 96\% &\textbf{28\%}  &  92\% & \textbf{36\%}   & 90\% & \textbf{48\%} \\
RTE   & 86\% &\textbf{56\%}  &  98\% & \textbf{54\%}   &  78\% & \textbf{66\%} \\
\bottomrule 
\end{tabular}
}
\label{table: sst2_robust}
\end{table}

\begin{table*}[h]
\caption{Case study: We demonstrate the examples where Shapley value succeeds but LOO fails in computing sign-consistent instance scores. 
[pos] means positive sentiment class, while [neg] means negative sentiment class. 
We report the mean and std of the scores scaled by a factor of $1\mathrm{e}{-3}$ for readability. The dataset size n=2000.}
\centering
\resizebox{1.0\linewidth}{!}{
\begin{tabular}{c|c|c|c|c|c}
\toprule
Text example & Label & Shap & Shapley value & LOO & LOO value\\
\midrule
confident filmmaking and a pair of fascinating performances & [pos]& 5+/0- & 0.45 $\pm$0.23 & 4+/1- & 0.23$\pm$1.52\\
seriously, rent the Disney version & [neg]& 5+/0- & 1.85$\pm$0.72  & 3+/2- & -0.69$\pm$2.36 \\
\begin{tabular}[c]{@{}l@{}} it's not going to be everyone's bag of popcorn, but it definitely gives you something to chew on’\end{tabular}& [pos]& 0+/5-& -1.91$\pm$0.57& 3+/2- &0.23$\pm$1.34\\
would fit chan like a \$ 99 bargain-basement special & [neg]& 0+/5-& -0.36$\pm$0.20& 2+/3- &0.00$\pm$1.45\\
 \bottomrule
\end{tabular}
}
\label{table: robust_case_study}
\end{table*}

\section{Applications of Instance Attribution}
\label{sec: application_instance_attribution}
In this section, we compare the quality of instance scores calculated using various baselines on the 4 aforementioned NLP datasets in Sec.~\ref{sec: experiment}. We adopt qualitative and quantitative evaluation criteria including \emph{test prediction explanation}, \emph{data removal}, \emph{data selection}, and \emph{wrong label detection}. 
The other baselines include the influence function~\citep{koh2017understanding}, TracIn~\citep{pruthi2020estimating}, and representer point~\citep{yeh2018representer}. 
For the influence function, we use a more accurate approximation with conjugate gradients~\citep{martens2010deep}, which is perceived as the exact influence function with enough iterations and often performs better than the most used approximation, LiSSA, as shown in App.~\ref{app: ablation_appro}. 
Due to the improved scalability, we adopt the modern practical prompt-based fine-tuning regime with sizable pre-trained models including BERT and RoBERTa~\citep{liu2019roberta}. More experimental set details are in App.~\ref{app: instance_attribution_setup}.

\begin{table*}[t]
\caption{Correctly predicted test samples with positive sentiment in SST-2 interpreted by different instance attribution approaches. [pos] means positive sentiment class, while [neg] means negative sentiment class. 
Words annotated in \textcolor{blue}{blue} indicate negative sentiment, while those in \textcolor{red}{red} suggest positive sentiment. These highlights are author-chosen for improved readability.
}
\centering
\resizebox{1.0\linewidth}{!}{
\begin{tabular}{c|cc|cc}
\toprule
\multicolumn{4}{c}{Test input: a quiet, pure, elliptical \textit{film}} & [pos]\\
\midrule
Attribution &\multicolumn{2}{|c|}{Most \textbf{helpful} training examples} & \multicolumn{2}{c}{Most \textbf{harmful} training examples}\\
\midrule
FreeShap & a \textcolor{red}{funny} little \textit{film} & [pos] & a particularly \textcolor{blue}{slanted} , *** */* \textit{fantasy} & [neg] \\
\midrule
Influence & goes down easy & [neg] & a guiltless \textit{film} & [pos]\\  
 \midrule
TracIn & \begin{tabular}[c]{@{}l@{}} the only young people who possibly will \\enjoy it  \end{tabular} & [neg] & \begin{tabular}[c]{@{}l@{}} earnhart 's \textit{film} is more about the \textcolor{red}{optimism} of a group of people who are \\struggling to give themselves a \textcolor{red}{better lot} in life than the ones \end{tabular} & [pos]\\
\midrule
Representer & a guiltless \textit{film} & [pos] & seagal & [neg]\\
\bottomrule 
\end{tabular}
}
\label{table: sst2-explanation-positive}
\end{table*}
\textbf{Test Prediction Explanation.}
To explain a test prediction, we analyze the helpful and harmful training examples categorized by the sign of the instance scores. 
Tab.~\ref{table: sst2-explanation-positive} shows the \emph{most} helpful/harmful examples (with the highest/lowest scores) of the instance attribution approaches. FreeShap's explanations align more closely with the test example syntactically, discussing related topics about films, while the explanations provided by other approaches are less related. Moreover, the helpful examples provided by FreeShap are semantically akin to the test example, whereas the identified harmful examples exhibit opposing semantics. Notably, by looking at more top-ranked examples, FreeShap provides consistently more meaningful explanations, as shown in Tab.~\ref{table: sst2-explanation-positive-full} in App.~\ref{app: explanations}. 
The advantages of FreeShap are likely due to its consideration of the subset's marginal contribution. 
When trained with smaller subsets, the semantically and syntactically relevant examples are likely to have more impact and score higher.
Additional case studies such as error explanations and attribution analyses are also in App.~\ref{app: explanations}.

\begin{figure*}[!htb]
\centering
\includegraphics[width=17cm]{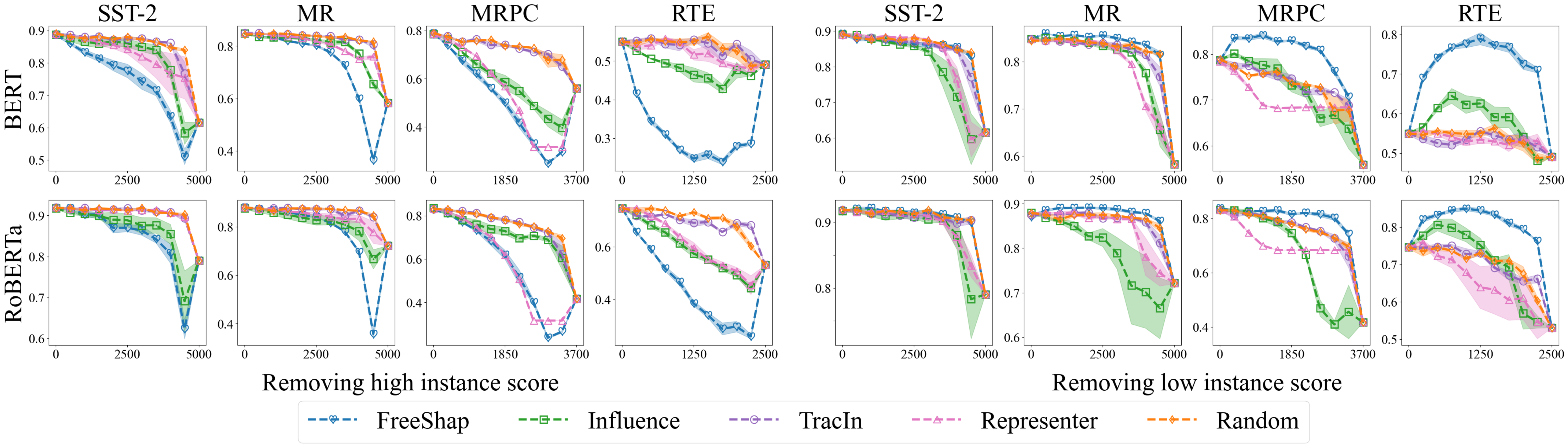}
\caption{\textbf{Data Removal}: The test accuracy on models retrained on subsets obtained by iteratively removing 10\% of the data, either from the highest or the lowest instance score. Faster degradation is preferable for high-score removals, while improvement or slower degradation is ideal for low-score removals. Overall, the scores from FreeShap are better correlated with test performance.
}
\label{fig: data_removal}
\end{figure*}

\textbf{Data Removal.} 
To quantitatively assess the correlation between the instance scores and the contributions to the test performance increase, we perform data removal experiments. This involves sequentially removing the data points 10\% at a time in \emph{order} from the training set, followed by evaluating the performance of the model retrained on the remaining data. The order is either from the highest to the lowest instance scores or vice versa, and we demonstrate for both, which is more comprehensive than the setting of~\citet{pezeshkpour-etal-2021-empirical} (i.e., only removing the highest score instances). 
As depicted in Fig.~\ref{fig: data_removal}, the left four columns show that when using the instance scores from FreeShap, the performance degrades comparably faster than other baselines when discarding the examples starting from the highest scores. Conversely, when discarding from the lowest scores, using FreeShap results in slower degradation and even improves the test performance in 3 out of the 4 datasets except for SST-2.
As other baselines are essentially based on the LOO framework, the consideration of more comprehensive marginal contributions in FreeShap leads to a more accurate estimation of the instance scores. 

\textbf{Data Selection.} With the effectiveness of FreeShap, we further conduct an ablation study to examine its potential in data selection settings. Specifically, we sequentially add the data points with the highest score (computed on the test set) and evaluate the generalized performance on a held-out set. 
We empirically evaluate the instance attribution approaches on the MR dataset with BERT model.
From Tab.~\ref{table: mr_addition_bert}, using FreeShap leads to faster performance improvements, validating its potential for data selection.
More results on other models and removal settings are in App.~\ref{app: data-selection}.

\begin{table}[h]
\caption{
\textbf{Data Selection:} Difference between the model performance of no training data (i.e., 0-shot) and the model performance when trained on a subset that contains data points with high instance scores. The subset size is given as a percentage of the full set. The higher the difference, the better the approach.
}
\centering
\resizebox{1.0\linewidth}{!}{
\begin{tabular}{c|c|c|c|c|c}
\toprule
Subset Size
&  2\% &  4\% &  6\% &  8\% &  10\% \\
\hline
FreeShap     & 0.1951 & \textbf{0.2111} & \textbf{0.2148} & \textbf{0.2167} & \textbf{0.2223}  \\ 
Influence   & 0.0272 & 0.0647 & 0.0393 & 0.0647 & 0.0750 \\
TracIn      & 0.1370 & 0.1904 & 0.2008 & 0.1548 & 0.1970 \\
Representer & 0.1182 & 0.1351 & 0.1388 & 0.0619 & 0.1238 \\
Random      & \textbf{0.1970} & 0.1951 & 0.2073 & 0.2017 & 0.1529 \\
\midrule
\end{tabular}
}
\label{table: mr_addition_bert}
\end{table}

\begin{figure}[!htb]
\centering
\includegraphics[width=8cm]{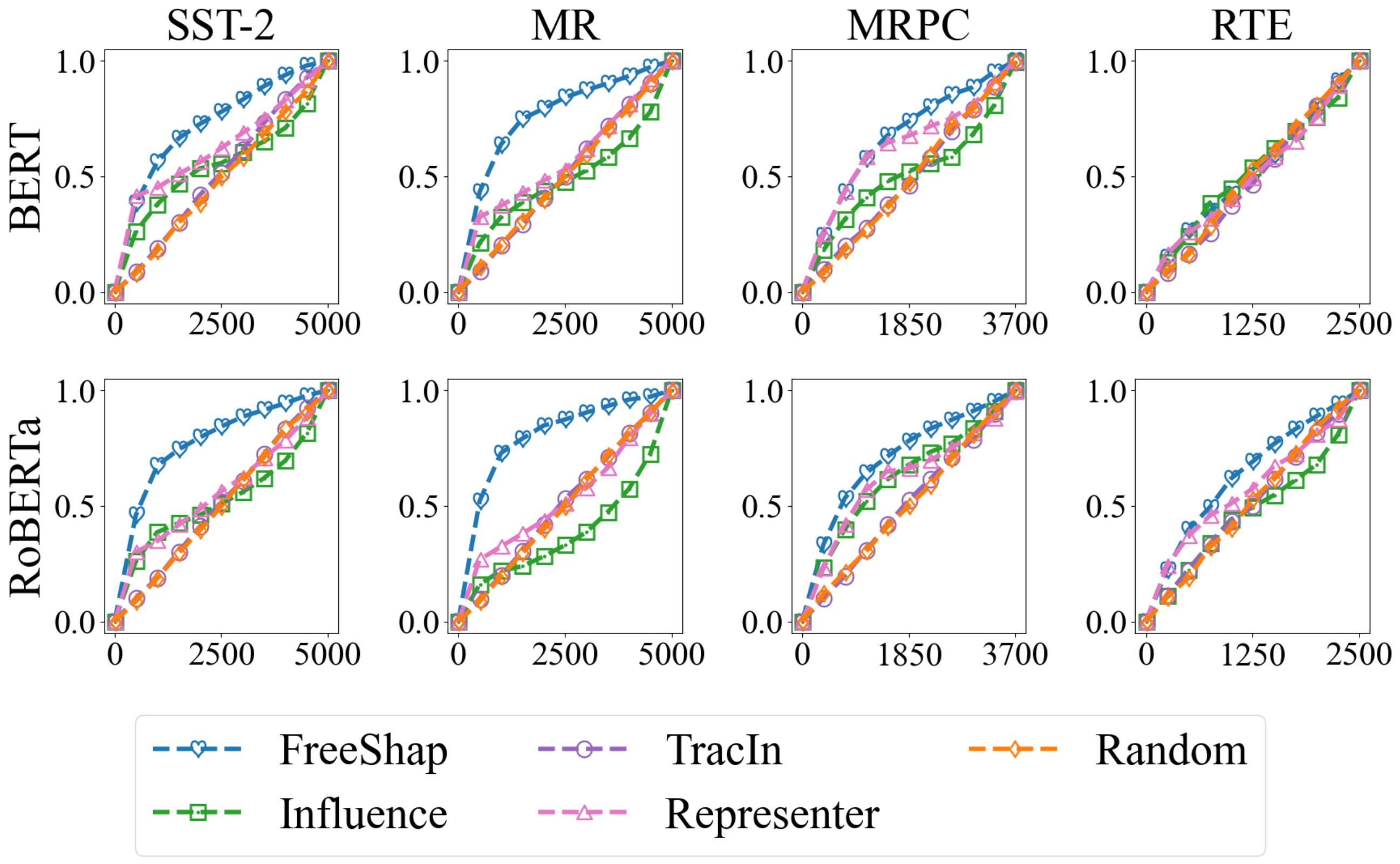}
\caption{\textbf{Wrong Label Detection}: 
It shows the detected percentage of poisoned data when inspecting data from lowest to highest instance score.
In most cases, FreeShap leads to the earliest identification of incorrectly labeled instances.}
\label{fig: data_poison}
\end{figure}

\textbf{Wrong Label Detection.}
Datasets can be mislabeled due to crowd-sourcing errors~\citep{frenay2013classification} or data tampering~\citep{steinhardt2017certified}, 
which may have detrimental effects on model training such as increased reliance on spurious correlations~\citep{qiao2024understanding}.
We compare the effectiveness of identifying mislabeled data using different instance attribution approaches.
We poison each training set by flipping 10\% of the labels, then automatically examine the poisoned data by reviewing the data points in order of their instance scores, starting from the lowest to the highest.
As shown in Fig.~\ref{fig: data_poison}, FreeShap is significantly more effective in identifying mislabeled data in 7 out of 8 experiments, while being comparable otherwise. 
The superiority of FreeShap in detecting mislabeled data is likely attributed to its comprehensive evaluation of the marginal contributions for the subsets as smaller data subsets are more vulnerable to the harmful effects of mislabeled data.

\begin{table*}[t]
\caption{A correctly predicted example with positive sentiment in SST-2 interpreted by different attribution approaches on prompt-based fine-tuned \textbf{Llama2} model. [pos] means positive sentiment class and predicted correctly, while [pos(neg)] means a positively labeled training example being wrongly predicted as negative.}
\vskip 0.15in
\centering
\resizebox{1.0\linewidth}{!}{
\begin{tabular}{c|cc|cc}
\toprule
\multicolumn{4}{c}{Test input: ... a \textcolor{red}{magnificent} \textit{drama} \textcolor{red}{well worth} tracking down . } & [pos]\\
\midrule
Attribution &\multicolumn{2}{|c|}{Most \textbf{helpful} training examples} & \multicolumn{2}{c}{Most \textbf{harmful} training examples}\\
\midrule
FreeShap & a \textcolor{red}{consummate} \textit{actor} incapable of being boring & [pos] &\begin{tabular}[c]{@{}l@{}} others will find their humor-seeking dollars \\best spent elsewhere.\end{tabular} & [neg] \\
\midrule 
Influence & most of it given to children & [pos(neg)] & \begin{tabular}[c]{@{}l@{}} 4ever has the same \textcolor{red}{sledgehammer} \\ \textcolor{red}{appeal} as pokemon \textit{videos} , but \end{tabular} & [pos(neg)]\\ 
 \midrule
TracIn & \textcolor{red}{superb} & [pos] & the book 's \textcolor{blue}{irreverent} energy , and scotches most & [neg]\\
\midrule
Representer & \begin{tabular}[c]{@{}l@{}} you 've endured a long workout without \\your pulse ever racing \end{tabular} & [pos(neg)] & \textcolor{blue}{self-deprecating, biting} and \textcolor{red}{witty} feature & [neg]\\
\bottomrule 
\end{tabular}
}
\label{table: sst2-explanation-positive-llama}
\end{table*}
\begin{figure*}[!htb]
\centering
\includegraphics[width=17cm]{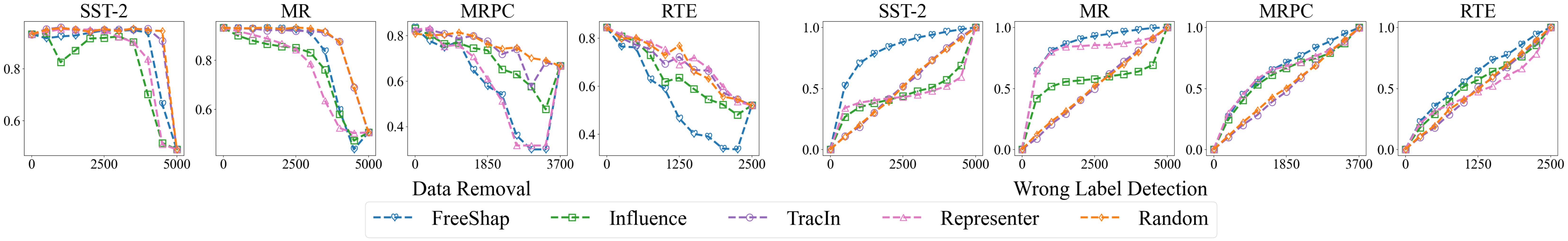}
\caption{\textbf{Data Removal and Wrong Label Detection for Llama2}: When removing data from the highest instance scores, the data points identified by FreeShap cause relatively faster degradation (more significant on the RTE dataset) and hence are more essential for the model. For wrong label detection, FreeShap identifies the incorrectly
 labeled instances earlier.}
\label{figure: data_removal_llm}
\end{figure*}

\textbf{Extension to LLMs.} In light of the recent breakthroughs in LLM~\citep{achiam2023gpt}, we extend our empirical study to assess the efficacy of FreeShap on Llama2~\citep{touvron2023llama2}, which is made possible by the scalability of FreeShap.  
We perform data removal and wrong label detection experiments on the prompt-based fine-tuning for Llama2 with LoRA~\citep{hu2022lora}. 
Details on experimental settings are in App.~\ref{app: supportiveness_llama_setup}. Since the influence function with conjugate gradients is not computationally feasible at the scale of the Llama2, we use the LiSSA approximation. 

We present the results for the test prediction explanation in Tab.~\ref{table: sst2-explanation-positive-llama}. FreeShap consistently provides more insightful explanations in terms of both syntactic similarity to the test example and semantic alignment. We also present the results for data removal (of high instance score) and wrong label detection in Fig.~\ref{figure: data_removal_llm}. 
The high-score examples identified by FreeShap have more impact on model performance, especially for RTE, and the low-score examples are more accurate in pinpointing mislabeled data.
We also demonstrate more case studies when FreeShap offers explanations, data removal of low instance score, time efficiency, and an extension to Llama2-13B in App~\ref{app: llama_instance_exp}.

\section{Related work}
We discuss related works on instance attribution and different existing notions of robustness. More discussions on Shapley value approximation and its usage in NLP explainability are in App.~\ref{app: related_work}.

\textbf{Instance attribution in NLP.}
Various methods have been developed to assess the influence of training instances on model predictions~\citep{pezeshkpour-etal-2021-empirical}. The influence function~\citep{koh2017understanding} identifies the impact of a training instance by approximating the LOO score. Its explainability and time complexity are subsequently improved by~\citet{pmlr-v108-barshan20a, grosse2023studying, guo-etal-2021-fastif, kwon2023datainf}. \citet{han-etal-2020-explaining} applies the influence function to the BERT model and shows its faithfulness and the ability to reveal the reasoning process. 
TracIn~\cite{pruthi2020estimating} quantifies the effect of each training example on a test prediction based on the overall loss reduction using the first-order Taylor approximation. Its NLP variant, TracIn-WE~\citep{yeh2022better}, exploits the word embedding layer for better instance scores. Representer point ~\citep{tsai2023sample, yeh2018representer} enables the model diagnosis for misclassified instances. These works all fall under the LOO scheme. In contrast, our work focuses on the Shapley value and we demonstrate its robustness and quality for data over the LOO methods.
Source attribution~\cite{wang2023wasa}, another attribution scheme that finds the sources of the contents present in a sentence generated from an LLM, is different from the instance attribution and hence is omitted from our discussion here.

\textbf{Robust Explanations.} Robustness determines the stability of explanations when inputs are slightly altered~\citep{subhash2022what}. The explanations are expected to remain similar when inputs are insignificantly modified. 
Robust explanations have been shown to increase user trust~\citep{NEURIPS2019_a7471fdc}. Robustness to three types of perturbations are discussed: general perturbations which modify the input randomly and locally~\citep{alvarez2018robustness, ijcai2020p417, tang-etal-2022-identifying}; adversarial perturbations that mislead explanations~\citep{Ghorbani_Abid_Zou_2019, ivankay2022fooling}; and group-based perturbations that change the membership of sensitive features in inputs, examining fairness~\citep{dai2021will}. The prior work of robustness focuses more on feature attributions and perturbation-based robustness, while we raise a complimentary notion of robustness, targeting instance attribution w.r.t. training data's distribution. The work of~\citet{lin2024dist} also studies the notion of distributional robustness in data valuation. However, the robustness result is w.r.t. validation distribution, hence different from our work.

\section{Conclusion}
We propose a new notion of robustness for instance attribution and demonstrate the superiority of Shapley value in achieving robustness compared to other methods. 
To mitigate the high computational cost of the Shapley value, we introduce FreeShap and empirically demonstrate its efficiency and effectiveness in providing better instance attribution in prompt-based fine-tuning for LMs and LLMs. 
Our approach facilitates improving the trustworthiness of the model and can inspire future research in model interpretation, data curation, and data selection. However, our study is limited to the NLP domain with the assumption of having white-box access to the models. Additionally, we concentrate on explaining predictions oriented toward classification. Extending FreeShap to text generation tasks is non-trivial due to their autoregressive nature, which is left for future work.

\newpage
\section*{Acknowledgement}
We would like to thank the anonymous reviewers and AC for the constructive and helpful feedback. This research/project is supported by the National Research Foundation, Singapore under its AI Singapore Programme (AISG Award No: AISG$2$-PhD/$2021$-$08$-$017$[T]).
This research is supported by the National Research Foundation Singapore and the Singapore Ministry of Digital Development and Innovation, National AI Group under the AI Visiting Professorship Programme (award number AIVP-$2024$-$001$). Jingtan Wang is supported by the Institute for Infocomm Research of Agency for Science, Technology and Research (A*STAR).

\section*{Impact Statement}
This paper presents work whose goal is to advance the field of Machine Learning. There are many potential societal consequences of our work, none which we feel must be specifically highlighted here.

\bibliography{references}
\bibliographystyle{icml2024}

%%%%%%%%%%%%%%%%%%%%%%%%%%%%%%%%%%%%%%%%%%%%%%%%%%%%%%%%%%%%%%%%%%%%%%%%%%%%%%%
%%%%%%%%%%%%%%%%%%%%%%%%%%%%%%%%%%%%%%%%%%%%%%%%%%%%%%%%%%%%%%%%%%%%%%%%%%%%%%%
% APPENDIX
%%%%%%%%%%%%%%%%%%%%%%%%%%%%%%%%%%%%%%%%%%%%%%%%%%%%%%%%%%%%%%%%%%%%%%%%%%%%%%%
%%%%%%%%%%%%%%%%%%%%%%%%%%%%%%%%%%%%%%%%%%%%%%%%%%%%%%%%%%%%%%%%%%%%%%%%%%%%%%%
\newpage
\appendix
\onecolumn

\section{Proof for the Main Paper}\label{appendix:proofs}
\subsection{Proof of Theorem~\ref{theorem:robust-sv-loo}}
\begin{proof}
When $z_i$ is a \textit{consistently helpful} point: 
$\mathbb{P}_{D_N \sim \mathcal{P}^{n-1}|z_i}\biggl( \text{sgn}\Bigl(g^{\text{Shap}}(z_i, D_T, D_N)\Bigl) \neq \text{sgn}^*(z_i) \biggl)$ is equivalent to
$\mathbb{P}_{D_N \sim \mathcal{P}^{n-1}|z_i}\biggl(g^{\text{Shap}}(z_i, D_T, D_N) < 0 \biggl)$.
\begin{equation}\label{eqn:chev-inequality}
\begin{aligned}
&\mathbb{P}_{D_N \sim \mathcal{P}^{n-1}|z_i}\Bigl(g^{\text{Shap}}(z_i, D_T, D_N) \le \mathbb{E}_{D_N \sim \mathcal{P}^{n-1}|z_i}\bigl(g^{\text{Shap}}(z_i, D_T, D_N)\bigl) - t \Bigl) \\
    & \le \mathbb{P}(\bigl|g^{\text{Shap}}(z_i, D_T, D_N)- \mathbb{E}\bigl(g^{\text{Shap}}(z_i, D_T, D_N)\bigl)\bigl| \ge t) \\
    & \le \frac{\text{Var}\bigl(g^{\text{Shap}}(z_i, D_T, D_N)\bigl)}{t^2}
\end{aligned}
\end{equation}
The second inequality of Eqn.~\eqref{eqn:chev-inequality} holds according to Chebyshev's inequality. 
\begin{equation}
\begin{aligned}
\label{eqn: variance}
\text{Var}\bigl(g^{\text{Shap}}(z_i, D_T, D_N)\bigl) &= \frac{1}{n^2}\Bigr[\sum_{k=0}^{n-1}\text{Var}\bigl(\Delta_{z_i}^{D_N}(k,D_T)\bigl) +\\
&\sum_{p,q \in \{0,\dots,n-1\}, p\neq q} \text{Cov}\bigl(\Delta_{z_i}^{D_N}(p,D_T), \Delta_{z_i}^{D_N}(q,D_T)\bigl)\Bigr] \\ 
& \le \frac{1}{n^2} \Bigr[\sum_{k=0}^{n-1}\text{Var}\bigl(\Delta_{z_i}^{D_N}(k,D_T)\bigl) +\\
&\sum_{p,q \in \{0,\dots,n-1\}, p\neq q} \frac{\text{Var}\bigl(\Delta_{z_i}^{D_N}(p,D_T)\bigl)+ \text{Var}\bigl(\Delta_{z_i}^{D_N}(q,D_T)\bigl)}{2} \Bigr]\\
&= \frac{1}{n^2} \Bigr[\sum_{k=0}^{n-1}\frac{\text{Var}(\Delta_{z_i}^{D_N}(k,D_T))+ \text{Var}\bigl(\Delta_{z_i}^{D_N}(k,D_T)\bigl)}{2} +\\
&\sum_{p,q \in \{0,\dots,n-1\}, p\neq q} \frac{\text{Var}(\Delta_{z_i}^{D_N}(p,D_T))+ \text{Var}\bigl(\Delta_{z_i}^{D_N}(q,D_T)\bigl)}{2} \Bigr]\\
& = \frac{1}{n^2} \Bigr[\sum_{p,q \in \{0,\dots,n-1\}} \frac{\text{Var}(\Delta_{z_i}^{D_N}(p,D_T))+ \text{Var}\bigl(\Delta_{z_i}^{D_N}(q,D_T)\bigl)}{2} \Bigr]\\
&= \frac{1}{n^2}\Bigr[n \sum_{p=0}^{n-1}\text{Var}\bigl(\Delta_{z_i}^{D_N}(p,D_T)\bigl)\Bigr]\\
&= \frac{1}{n}\sum_{k=0}^{n-1} \delta_k
\end{aligned}
\end{equation}
The first inequality of Eqn.~\eqref{eqn: variance} holds because:
\begin{equation}
\begin{aligned}
\text{Cov}(\Delta_{z_i}^{D_N}(p,D_T), \Delta_{z_i}^{D_N}(q,D_T)) 
&= \mathbb{E} \Bigr[\bigl( \Delta_{z_i}^{D_N}(p,D_T) - \tau_p \bigl) \bigl( \Delta_{z_i}^{D_N}(q,D_T) - \tau_q \bigl) \Bigr]\\
& \le \mathbb{E} \Bigr[ \frac{ {\bigl(  \Delta_{z_i}^{D_N}(p,D_T) - \tau_p \bigl)}^2 + {\bigl(  \Delta_{z_i}^{D_N}(q,D_T) - \tau_q \bigl)}^2}{ 2 } \Bigr]\\
&= \frac{ \text{Var}\bigl(  \Delta_{z_i}^{D_N}(p,D_T)\bigl) + \text{Var}\bigl(  \Delta_{z_i}^{D_N}(q,D_T) \bigl)}{ 2 } 
\end{aligned}
\end{equation}
Substituting $t$ in Eqn.~\eqref{eqn:chev-inequality} to $\mathbb{E}_{D_N \sim \mathcal{P}^{n-1}|z_i}\bigl(g^{\text{Shap}}(z_i, D_T, D_N)\bigl)$, we have:
\begin{equation*}
\begin{aligned}
    \mathbb{P}_{D_N \sim \mathcal{P}^{n-1}|z_i}\Bigl(g^{\text{Shap}}(z_i, D_T, D_N) < 0 \Bigl) 
    & \leq
    \mathbb{P}_{D_N \sim \mathcal{P}^{n-1}|z_i}\Bigl(g^{\text{Shap}}(z_i, D_T, D_N) \leq 0 \Bigl) \\
    & \leq \frac{n^{-1} \sum_{k=0}^{n-1} \delta_k}{\Bigl(\mathbb{E}_{D_N \sim \mathcal{P}^{n-1}|z_i}\bigl(g^{\text{Shap}}(z_i, D_T, D_N)\bigl)\Bigl)^2} \\
    & \leq \frac{n^{-1} \sum_{k=0}^{n-1} \delta_k}{(n^{-1} \sum_{k=0}^{n-1} \tau_k)^2} \ .
\end{aligned}
\end{equation*}
Similarly, we have:
\begin{equation*}
    \mathbb{P}_{D_N \sim \mathcal{P}^{n-1}|z_i}\Bigl(g^{\text{LOO}}(z_i, D_T, D_N) < 0 \Bigl) 
    \le \frac{\delta_{n-1}}{\tau_{n-1}^2} \ .
\end{equation*}

Consider two edge cases when $\mathbb{E}_{D_N \sim \mathcal{P}^{n-1}|z_i}\bigl(g^{\text{LOO}}(z_i, D_T, D_N)\bigl) = 0$ while the expectation of Shapley is not $0$ and when $\mathbb{E}_{D_N \sim \mathcal{P}^{n-1}|z_i}\bigl(g^{\text{Shap}}(z_i, D_T, D_N)\bigl) = 0$ while the expectation of LOO is $0$: For the first case, the upper bound for Shapley value is still well defined, and the upper bound for LOO is infinite. For the latter case, both upper bounds for Shapley and LOO are infinite. Even if the bounds derived from Chebyshev's inequality are not well-defined, the bounds still apply to both Shapley and LOO. The subsequent bound comparison is valid as well.

When $z_i$ is an \textit{consistently harmful} data point: 
$\mathbb{P}_{D_N \sim \mathcal{P}^{n-1}|z_i}\biggl( \text{sgn}\Bigl(g^{\text{Shap}}(z_i, D_T, D_N)\Bigl) \neq \text{sgn}^*(z_i) \biggl)$ is equivalent to
$\mathbb{P}_{D_N \sim \mathcal{P}^{n-1}|z_i}\biggl(g^{\text{Shap}}(z_i, D_T, D_N) \geq 0 \biggl)$

\begin{equation}\label{eqn:chev-inequality-2}
\begin{aligned}
&\mathbb{P}_{D_N \sim \mathcal{P}^{n-1}|z_i}\Bigl(g^{\text{Shap}}(z_i, D_T, D_N) \geq \mathbb{E}_{D_N \sim \mathcal{P}^{n-1}|z_i}\bigl(g^{\text{Shap}}(z_i, D_T, D_N)\bigl) + t \Bigl) \\
    & \le \mathbb{P}(\bigl|g^{\text{Shap}}(z_i, D_T, D_N)- \mathbb{E}\bigl(g^{\text{Shap}}(z_i, D_T, D_N)\bigl)\bigl| \ge t) \\
    & \le \frac{\text{Var}\bigl(g^{\text{Shap}}(z_i, D_T, D_N)\bigl)}{t^2}.
\end{aligned}
\end{equation}
The second inequality of Eqn.~\ref{eqn:chev-inequality-2} holds according to Chebyshev's inequality. Substituting $t$ in Eqn.~\eqref{eqn:chev-inequality} to $-\mathbb{E}_{D_N \sim \mathcal{P}^{n-1}|z_i}\bigl(g^{\text{Shap}}(z_i, D_T, D_N)\bigl)$, we have:
\begin{equation}
\begin{aligned}
    \mathbb{P}_{D_N \sim \mathcal{P}^{n-1}|z_i}\Bigl(g^{\text{Shap}}(z_i, D_T, D_N) \geq 0 \Bigl)
    & \le \frac{n^{-1} \sum_{k=0}^{n-1} \delta_k}{\Bigl(- \mathbb{E}_{D_N \sim \mathcal{P}^{n-1}|z_i}\bigl(g^{\text{Shap}}(z_i, D_T, D_N)\bigl)\Bigl)^2} \\
    & \le \frac{n^{-1} \sum_{k=0}^{n-1} \delta_k}{(n^{-1} \sum_{k=0}^{n-1} \tau_k)^2} \ .
\end{aligned}
\end{equation}

Similarly, we have:
\begin{equation}
    \mathbb{P}_{D_N \sim \mathcal{P}^{n-1}|z_i}\Bigl(g^{\text{LOO}}(z_i, D_T, D_N) \geq 0 \Bigl) 
    \le \frac{\delta_{n-1}}{\tau_{n-1}^2} \ .
\end{equation}

\end{proof}

\section{Improving the Efficiency of FreeShap}
\label{app: time_TFSV}
We adopt two tricks when using the eNTK to emulate fine-tuning: one for efficient kernel approximation and the other for efficient kernel regression. 

\subsection{Efficient Kernel Approximation}
For eNTK matrix $K(X_N, X_N)$ where $N \coloneq \{1, \dots, n\}$, the matrix belongs to $\mathbb{R}^{nC \times nC}$. The complete eNTK can be approximated by $I_C \otimes K_c$, where $K_c \in \mathbb{R}^{n \times n}$ is a kernel matrix concerning the output logit of class $c$, and $I_C$ is the $C \times C$ identity matrix. This approximation allows kernel regression for $K(X_N, X_N)$ to be split into $C$ separate kernel regression tasks, each concerning $K_c$. If the model has randomly initialized output layers, ~\citet{pmlr-v162-wei22a} further approximates the kernel matrix by $I_C \otimes K_0$ where $K_0$ is the eNTK concerning a random output logit. As we adopt prompt-based fine-tuning, our output layer is based on the embedding layer which is not randomly initialized. Thus, we experiment with computing only one $K_0$ (which is more time efficient) as well as computing different $K_c$. 
Following the setting in Sec~\ref{sec: corr_exp} using SST-2 dataset, we compare the correlation for one seed with MC-Shapley when using each class's $K_c$ for kernel regression and using $K_0$ (the eNTK concerning a random output logit). It shows a similar correlation, the former is $0.69$ while the latter is $0.71$. Considering the similar approximation performance and computing only one $K_0$ shows an even better approximation, we use the same $K_0$ which is the eNTK concerning one of the class's outputs to conduct $C$ separate kernel regression tasks in all our experiments for computational efficiency.

\subsection{Efficient Kernel Regression}
When generalizing to the dataset with a larger size, the computation of the kernel regression can be more computationally expensive as it requires the inversion of the matrix $K(X_S, X_S)$, as shown in Eqn.~\eqref{eq:kernel_regression}. Instead of inverting the full matrix from scratch, we use a blockwise inversion (BI) approach: we make use of the last step's inverse matrix and reduce the average time complexity of kernel regression from $O(n^3)$ to $O(n^2)$. Specifically, when computing ${K(X_S, X_S)}^{-1}$, we partition this inverse into four blocks: 
\begin{equation}
{K(X_S, X_S)}^{-1} = \begin{bmatrix}
A & B \\
C & D
\end{bmatrix}^{-1} = \begin{bmatrix}
A^{-1} + A^{-1}B(D - CA^{-1}B)^{-1}CA^{-1} & -A^{-1}B(D - CA^{-1}B)^{-1} \\
-(D - CA^{-1}B)^{-1}CA^{-1} & (D - CA^{-1}B)^{-1}
\end{bmatrix}
\end{equation}
Technically, BI is a dynamic programming approach that efficiently computes the exact inverse. Nonetheless, to check the possible numerical instability in practice, we still empirically compare the Shapley value of the SST-2 dataset computed using the BI and without the BI (the exact inverse). We compute the Shapley value for 5k SST-2 points. For each test point, there is a (5000,1) vector, denoting the instance score for each training point. For each test point, there is a correlation value between computing using the BI and without the BI. We average all test points' correlation values and get a $0.9984$ Pearson correlation and $0.9963$ Spearman correlation. The decent correlation demonstrates that kernel regression using BI offers an efficient yet accurate computation of the inverse. This justifies our later usage of BI when the dataset is relatively larger in Sec.~\ref{sec: application_instance_attribution}.

\section{More Related Works}
\label{app: related_work}

\textbf{Approximating the Shapley Value.} 
Numerous works have focused on improving the scalability of the Shapley value by reducing the number of computations of the subset's utility, such as using Monte-Carlo sampling~\citep{pmlr-v97-ghorbani19c}, adopting group testing for fewer subsets utility computation~\citep{pmlr-v89-jia19a}, or updating multiple estimates simultaneously with one time of utility computation~\citep{kolpaczki2023approximating}. 
On the other hand, some works aim to reduce the computation of the utility function as ours. Gradient Shapley~\citep{pmlr-v97-ghorbani19c} approximates the full training with just one epoch, while All-S Influence Shapley and Largest-S Influence Shapley~\citep{pmlr-v89-jia19a} leverage the influence function to approximate the marginal contribution. The influence-function-based approximations require repeated Hessian calculations or consider only a single subset, losing some of Shapley value's essential properties. Consequently, they are not selected as baselines in our analysis. We show that our work provides a better approximation to the original Shapley values with comparable efficiency. 
The work of~\citet{pmlr-v162-wu22j} proposes a domain-aware approximation of generalization error of neural network model based on NTK and distributional distance. However, it is unclear how it can be used to approximate validation accuracy in the setting of prompt-based fine-tuning.
Other works extract fixed representations from the last hidden state and then conduct KNN~\cite{9578820} or last-layer fine-tuning \citep{schoch-etal-2023-data}. We did not make further comparisons with the previous two works as they focus solely on using representations and fine-tuning the last hidden layer, which \citet{yeh2022better} suggest can lead to suboptimal results for language models. In contrast, we aim to resemble a more comprehensive fine-tuning phase. FreeShap uses validation accuracy as the utility function to define Shapley value and more discussion on other utility functions can be found in~\citet{wu2024data}.

\textbf{Shapley Value in NLP Explainability.}
Shapley value has been employed for feature-attribution explanations~\citep{10.5555/3295222.3295230, mosca-etal-2022-shap}, treating each word as a player in a game, with the model's performance as the outcome. In terms of NLP, this approach has been adapted for text classification, identifying keywords or phrases influencing classification decisions. Various strategies for efficient Shapley value approximation have been developed, such as KernelSHAP, LinearSHAP, and DeepSHAP~\citep{10.5555/3295222.3295230}, targeting different model types. However, these methods oversimplify text data by not accounting for word interactions and context. HEDGE~\citep{chen-etal-2020-generating} offers hierarchical explanations, considering phrases and multiple relevance levels. In contrast to these feature-attribution applications, recent works have started using Shapley value for instance scores in NLP~\citep{schoch-etal-2023-data}, focusing on selecting training points with high Shapley values. Our work differs from these in both purpose and technique, aiming at robust instance attribution and utilizing eNTK to resemble fine-tuning.

\section{Detailed Experiment Setup}

\subsection{Detailed Experiment Setup for Sec.~\ref{sec: experiment}}
\label{app: robust_hyperparameter}
In terms of the model, we adopt the BERT~\cite{devlin-etal-2019-bert}, comprised of 12 layers of encoders. To address time complexity, we freeze the first eight layers of encoders and fine-tune the remaining layers. We adopt Adam optimizer during the training process. We disable the dropout in the BERT model to reduce the other uncertainty. We also disable the data shuffling when feeding data to the model.
In terms of implementation, we adopt the huggingface transformers package to implement BERT's fine-tuning.
In terms of hyperparameters, we use a learning rate of $1\text{e}{-5}$ (G-Shapley has a $2\text{e}{-5}$ learning rate), batch size $16$, and $10$ epochs. The max sequence length set for SST-2 dataset and MR dataset is $64$, for MRPC is $128$, and for RTE is $256$. The prompts for each dataset are listed in Tab.~\ref{table: prompt}, which is the same as ~\citet{pmlr-v202-malladi23a}'s. We adopt $200$ Monte-Carlo iterations for experiments in this section.
In terms of eNTK, we adopt the standard neural tangent kernel described in the main paper (Sec.~\ref{sec: ntk}), different from the suggested signGD kernel~\citep{pmlr-v202-malladi23a}. The analysis for the signGD kernel applies only to ``early-stage'' training with Adam, and previous empirical work is limited to few-shot classification. We experiment with both kernels: We compare the correlation between MC-Shapley and FreeShap when using the standard neural tangent kernel or signGD kernel and for seed 2023 on four datasets, and find that the standard neural tangent kernel overall has a relatively higher correlation compared with the signGD kernel. Therefore, we proceed with the standard neural tangent kernel. 
In terms of hardware, fine-tuning and eNTK precomputation are conducted on L40 GPUs. Notably, kernel regression can be efficiently computed with just CPU. 

\begin{figure}[h]
\begin{center}
\centerline{\includegraphics[width=10cm]{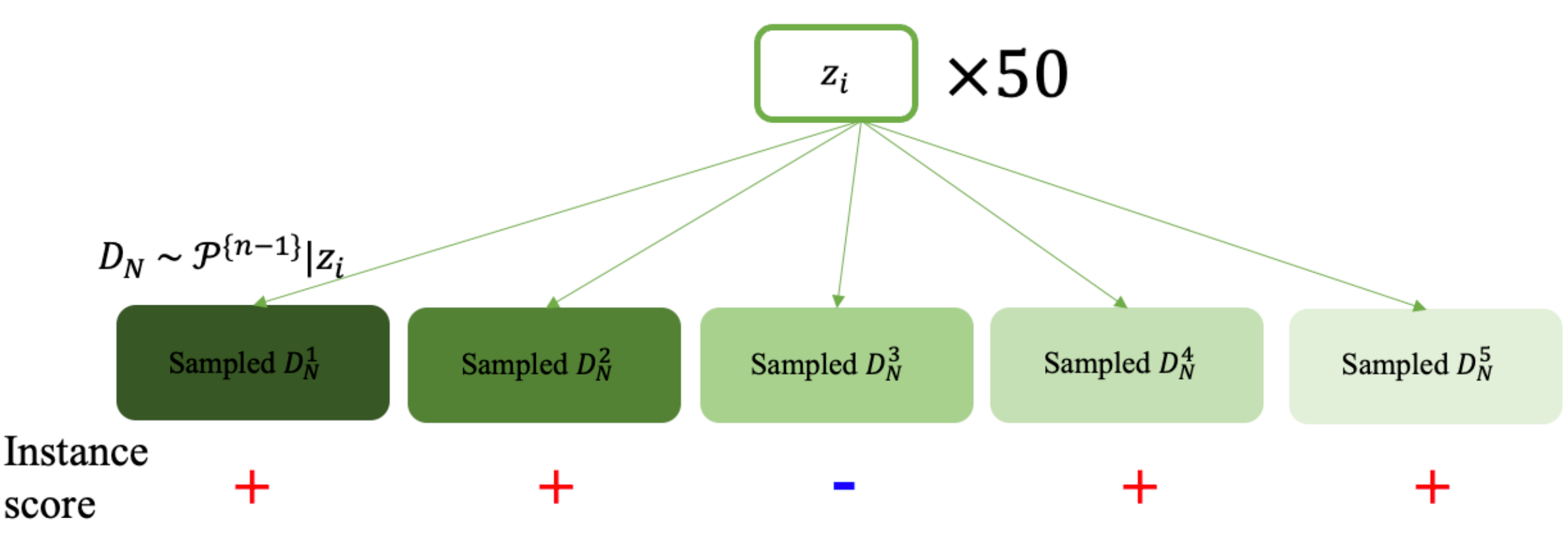}}
\caption{Experiment setup for robustness comparison. }
\label{fig: exp_setup}
\end{center}
\end{figure}

\begin{table}[h]
\centering
\caption{Prompts used in experiments. [x] in the single sentence task means the training input, while [x0] and [x1] are two sentences in the sentence pair task. The prompts for SST-2 and MR are the same for encoder-only or decoder-only models. }
\begin{tabular}{c|c|c|c}
\toprule
Dataset & Type & Prompts & Task-specific term\\
\midrule
\multicolumn{3}{c}{encoder-only model} \\
\midrule
SST-2    & sentiment & [x] It was mask.      & {terrible, great}                \\ 
MR     &  sentiment & [x] It was mask.      & {terrible, great}                \\ 
MRPC    & paraphrase   & [x0] mask, [x1]      & {No, Yes}                        \\ 
RTE     & NLI     & [x0]? mask, [x1]      & {No, Yes}                        \\ 
\midrule
\multicolumn{3}{c}{decoder-only model} \\
\midrule
MRPC    & paraphrase & [x0] Question: [x1] Yes or No? mask   & {No, Yes}  \\ 
RTE     & NLI  & [x0]? [x1] Entailment or not? mask   & {No, Yes}  \\ 
\bottomrule 
\end{tabular}
\label{table: prompt}
\end{table}

\newpage 

\subsection{Detailed Experiment Setup for Sec.~\ref{sec: application_instance_attribution}}
\label{app: instance_attribution_setup}
In terms of the model, we adopt BERT and RoBERTa~\citep{liu2019roberta}. The practical implementation and hyperparameters for both models are the same as the setup described in App.~\ref{app: robust_hyperparameter}. One difference is that we adopt TMC sampling~\cite{pmlr-v97-ghorbani19c} here instead of MC sampling for FreeShap, which has a performance tolerance of $0.05$ and $200$ Monte-Carlo iterations. Another difference is that we adopt BI as described in App.~\ref{app: time_TFSV} for the experiments in this section as the size of the dataset is larger. \\
In terms of other instance attributions: for the influence function, the damping factor is $0.003$, while the maximum iterations are $1000$; for TracIn, the checkpoint step is $3$; for representer point, the $\lambda$ is $1\text{e}{-4}$. Influence function and TracIn are computed on unfrozen parameters. 

\subsection{Detailed Experiment Setup for Llama2}
\label{app: supportiveness_llama_setup}

In terms of the model, we adopt Llama2-7B, comprised of 32 layers of decoders. We do not freeze the decoders as we adopt LoRA for efficient fine-tuning. In the LoRA experiment setup, the rank is $16$ while the alpha value is also $16$. Targeted modules for LoRA adjustments include query, key, value, output, gate, down, and up projections in the transformer's attention mechanism. We adopt no bias and dropout. Additionally, the final predictive layer is fine-tuned. 
In terms of hyperparameters, we use a learning rate of $1\text{e}{-5}$ or $1\text{e}{-6}$ and $5$ epochs. 
The batch size for the SST-2 dataset and MR dataset is $4$, and for MRPC and RTE is $2$. For Llama2, a decoder-only model, slightly different prompts are adopted. The prompts employed for Llama2 are similar to those used in transformer structure models in \citet{du-etal-2022-glm}. The prompts for each dataset for Llama2 are also listed in Tab.~\ref{table: prompt}. The SST-2 dataset and MR dataset share similar prompts as the previous setting so it is not repeated showing. 
The setup for FreeShap is the same as the setup described in App.~\ref{app: instance_attribution_setup}.
In terms of other instance attributions: for the influence function, the damping factor is $0.003$, repeat time is $1$, depth is $2500$ and the scaling factor is $1e4$; for TracIn, the checkpoint step is $1$; for representer point, the $\lambda$ is $1\mathrm{e}{-4}$. Influence function and TracIn are computed on unfrozen parameters.

\section{More Experimental Results}
\subsection{FreeShap Approximates the Shapley Value Well}
\label{app: more_result_corr}
Tab.~\ref{table:shapley-corr-1000} compliments the results of Tab.~\ref{table:shapley-corr}, further demonstrating that our approach is more effective in approximating MC-Shapley under different sizes of the dataset. 
Additionally, comparing Tab.~\ref{table:shapley-corr} and Tab.~\ref{table:shapley-corr-1000}, we also observe that when the dataset size is 500, both FreeShap and G-Shapley have a generally better approximation. Shapley values in size of 500 points and 1000 points share similar variance as they use the same number of Monte-Carlo iterations; while Shapley values in size of 500 points have a larger magnitude, and hence they may fluctuate less and better at maintaining the correlation. 

\begin{table}[htb]
\caption{Correlation between MC-Shapley when the dataset size is 1k. Each correlation is computed using three trials. The higher the better and FreeShap consistently has a higher correlation.}
\vskip 0.15in
\centering
\resizebox{0.5\linewidth}{!}{
\begin{tabular}{c|cc|cc}
\toprule
\multirow{2}{*}{Dataset} & \multicolumn{2}{c|}{FreeShap} & \multicolumn{2}{c}{G-Shapley}\\
 &  pearson & spearman & pearson & spearman \\
\midrule
\multicolumn{5}{c}{Single Sentence Task}\\
\midrule
SST-2  & \textbf{0.65$\pm$0.06}  &  \textbf{0.50$\pm$0.01}  &  0.35$\pm$0.08  &  0.39$\pm$0.16 \\ 
MR    &  \textbf{0.55$\pm$0.08}  &  \textbf{0.41$\pm$0.06}  &  0.17$\pm$0.07  &  0.17$\pm$0.04 \\
\midrule
\multicolumn{5}{c}{Sentence Pair Task}\\
\midrule
MRPC  & \textbf{0.79$\pm$0.05}  &  \textbf{0.75$\pm$0.03}  &  0.51$\pm$0.06  &  0.42$\pm$0.02 \\ 
RTE   & \textbf{0.61$\pm$0.10}  &  \textbf{0.59$\pm$0.13}   &  0.16$\pm$0.02  &  0.24$\pm$0.07 \\ 
\bottomrule 
\end{tabular}
}
\label{table:shapley-corr-1000}
\end{table}

\subsection{The Shapley Value Is More Robust than LOO}
\label{app: robust_result}

Table~\ref{table: sst2_robust_full} is a more comprehensive version of Tab.~\ref{table: sst2_robust}, demonstrating the Shapley value is more robust in instance scores compared to LOO. Across various training set sizes, the Shapley value generally exhibits fewer non-robust occurrences. Note that for the RTE dataset, the maximum size of the training data is 2,490. Therefore, the column $|D_N|=3000$ represents the scenario where $|D_N|$ is at its maximum.

Table~\ref{table: robust_case_study_appendix} presents more exemplary examples across various dataset sizes, where the Shapley value consistently has sign-consistent scores but LOO fails. This mirrors the main paper's findings: the Shapley value produces sign-consistent scores for training instances that resonate with human intuition, where negative scores signal confusing examples, and non-negative scores indicate more straightforward and stronger polarized examples. In the main paper, we have discussed an example with a negative instance score, while here we use ``looks great, has solid acting and a neat premise'' when dataset size is $3000$ as a case study. This example, with its clear positive sentiment, demonstrates that including clear-polarity training points can enhance test performance. The Shapley value consistently assigns a positive score to this example. However, LOO may incorrectly label this example as harmful, leading to potential misunderstandings about its influence or underrepresentation in future data collection and curation.

\begin{table}[htb]
\caption{A complete version of the empirical comparison of the $\beta$-robustness between the Shapley value and LOO by the percentage of non-robust occurrences when the complementary dataset is resampled. The lower, the better. }
\centering
\vskip 0.15in
\begin{tabular}{c|cc|cc|cc|cc|cc}
\toprule
\multirow[t]{2}{*}{size} & \multicolumn{2}{c|}{$|D_N|=300$}&  \multicolumn{2}{c|}{$|D_N|=500$} &  \multicolumn{2}{c|}{$|D_N|=1000$} & \multicolumn{2}{c|}{$|D_N|=2000$} & \multicolumn{2}{c}{$|D_N|=3000$}\\
dataset & LOO&Shapley  &  LOO&Shapley  &  LOO&Shapley  &  LOO&Shapley  &  LOO& Shapley\\
\midrule
SST-2  & \textbf{44\%} &56\%&  86\%&\textbf{30\%}  &  \textbf{38\%}&50\%  & 90\%&\textbf{46\%}  &  100\% &\textbf{52\%}  \\ 
MR  & 44\% &\textbf{24\%}&  98\%&\textbf{18\%}  &  100\%&\textbf{28\%}   & 100\%&\textbf{44\%}  &  98\% &\textbf{36\%}  \\
MRPC  & 100\% &\textbf{26\%}&  96\%&\textbf{28\%}  &  92\%&\textbf{36\%}   & 90\% &\textbf{48\%}  &  98\% &\textbf{4\%}  \\
RTE   & 98\% &\textbf{58\%}&  86\%&\textbf{56\%}  &  98\%&\textbf{54\%}   & 78\%&\textbf{66}\%  &  100\% &\textbf{48\%}  \\
\bottomrule 
\end{tabular}
\label{table: sst2_robust_full}
\end{table}

\begin{table}[htb]
\caption{
Case study: we demonstrate the examples with consistent instance scores computed by Shapley value and sign-inconsistent instance scores computed by LOO. [pos] means positive sentiment class, while [neg] means negative sentiment class. We report the mean and std of the scores scaled by a factor of $1\mathrm{e}{-3}$ for readability. The dataset size n=1000, 2000, 3000. }
\vskip 0.15in
\centering
\resizebox{0.9\linewidth}{!}{
\begin{tabular}{c|c|c|c|c|c}
\toprule
Text example & Label & Shap & Shapley value & LOO & LOO value\\
\midrule
\multicolumn{6}{c}{1000 points}\\
\midrule
prefer to keep on watching & [pos] & 0+/5- & -1.01$\pm$0.26  & 4+/1- & 0.00$\pm$1.26 \\
 \begin{tabular}[c]{@{}l@{}} it seems impossible that an epic four-hour indian musical \\about a cricket game could be this good\end{tabular}& [pos]& 0+/5-& -1.09$\pm$0.29& 3+/2- &0.69$\pm$2.96\\
 \midrule
\multicolumn{6}{c}{2000 points}\\
\midrule
confident filmmaking and a pair of fascinating performances & [pos]& 5+/0- & 0.45 $\pm$0.24 & 4+/1- & 0.23$\pm$1.52\\
seriously, rent the Disney version & [neg]& 5+/0- & 1.85$\pm$0.72  & 3+/2- & -0.69$\pm$2.36 \\
\begin{tabular}[c]{@{}l@{}} it's not going to be everyone's bag of popcorn, but it definitely \\ gives you something to chew on’\end{tabular}& [pos]& 0+/5-& -1.91$\pm$0.57& 3+/2- &0.23$\pm$1.34\\
would fit chan like a \$ 99 bargain-basement special & [neg]& 0+/5-& -0.36$\pm$0.20& 2+/3- &0.00$\pm$1.45\\
\midrule
\multicolumn{6}{c}{3000 points}\\
\midrule
looks great, has solid acting and a neat premise.& [pos]& 5+/0- & 0.58 $\pm$0.40& 3+/2- & 0.46$\pm$1.72\\
seriously, rent the Disney version & [neg]& 5+/0- & 1.17$\pm$0.60  & 2+/3- & -0.69$\pm$3.04 \\
 \begin{tabular}[c]{@{}l@{}} it's not going to be everyone's bag of popcorn, but it definitely \\gives you something to chew on ’\end{tabular}& [pos]& 0+/5-& -1.27$\pm$0.68& 2+/3- &0.00$\pm$2.71\\
they shovel into their mental gullets to simulate sustenance & [neg]& 0+/5-& -0.28$\pm$0.14& 3+/2- &0.23$\pm$2.22\\
 \bottomrule
\end{tabular}
}
\label{table: robust_case_study_appendix}
\end{table}

\newpage

\section{More Results for Applications of Instance Attributions} 
\subsection{More Test Prediction Explanations}

For the tables in this section: [pos] means positive sentiment class, while [neg] means negative sentiment class. Each training example shown here is predicted correctly so we do not include its prediction result. Words annotated in \textcolor{blue}{blue} indicate negative sentiment, while those in \textcolor{red}{red} suggest positive sentiment. These highlights are author-chosen for improved readability. The dataset used in this section is SST-2. 

We provide a more complete example for the \textbf{most} helpful/harmful examples (with
the highest/lowest scores) of the correct prediction in Tab.~\ref{table: sst2-explanation-positive-full}. 
The Shapley value yields consistently more insightful explanations for the top three most significant training examples, both in terms of syntactic similarity to the test example and semantic alignment. Specifically, helpful examples share identical semantics with the test example, whereas harmful examples demonstrate contrasting semantics. Other instance attribution methods might offer a film-related example in their explanations but fail to provide consistently meaningful explanations. 

We further list the explanations for an incorrect prediction in Tab.~\ref{table: sst2-explanation-negative-full}. 
The helpful examples provided by the Shapley value are syntactically similar to the test example in that they contain a mix of words with positive and negative connotations. This mixture could potentially confuse a deep learning model. The representer point also provides a meaningful explanation, ``peril'', a negative sentiment yet labeled under the positive class. This aligns with the findings of ~\citet{yeh2018representer} that representer points can offer insights into the neural network's tendency to predict the positive sentiment for this test example. This tendency stems from its reliance on examples that, despite conveying a negative sentiment, are labeled within the positive class.

Analyzing training data with high instance scores relative to the test set, as shown in Table~\ref{table: sst2-explanation-global}, guides new data collection by targeting similar profiles. Specifically, the Shapley value indicates that adding examples with longer sequences and clear polarity could significantly enhance model performance. Conversely, the model struggles with sarcasm, as evident from the most harmful examples. The influence function effectively identifies examples with incorrect labels. Interestingly, the most supportive example per the influence function aligns with the representer point but lacks clear positive polarity, with the term "odd" suggesting uniqueness rather than positivity. This example is assigned a negative score by Shapley, suggesting it may hinder performance in smaller subsets. Additionally, TracIn may be biased towards the same training examples when they have large gradients.

\label{app: explanations}
\begin{table}[htb]
\caption{A correctly predicted example with positive sentiment in SST-2 interpreted by different instance attribution approaches.
}
\vskip 0.15in
\centering
\resizebox{1.0\linewidth}{!}{
\begin{tabular}{c|cc|cc}
\toprule
\multicolumn{4}{c}{Test input: a quiet, pure, elliptical \textit{film}} & [pos]\\
\midrule
Attribution &\multicolumn{2}{|c|}{Most \textbf{helpful} training examples} & \multicolumn{2}{c}{Most \textbf{harmful} training examples}\\
\midrule
FreeShap & a \textcolor{red}{funny} little \textit{film} & [pos] & a particularly \textcolor{blue}{slanted} , *** */* \textit{fantasy} & [neg] \\
& \textcolor{red}{pretty good} little \textit{movie}  & [pos] &  \textcolor{blue}{overly-familiar} and \textcolor{blue}{poorly-constructed} \textit{comedy} & [neg]\\
& a \textcolor{red}{well-put-together} \textit{piece}  & [pos] &  a \textcolor{blue}{shaky, uncertain} \textit{film} & [neg]\\
\midrule 
Influence & goes down easy & [neg] & a guiltless \textit{film} & [pos]\\ 
& is like nothing we westerners have seen before & [pos]  & money-oriented & [neg]\\ 
& \begin{tabular}[c]{@{}l@{}}\textcolor{blue}{retread}, with the emphasis on \textcolor{red}{self-empowering} \\schmaltz and big-wave surfing that gives \textit{pic} \\its title an \textcolor{blue}{afterthought}\end{tabular} & [neg]  & is not as well-conceived as either of those \textit{films} & [neg]\\
 \midrule
TracIn & \begin{tabular}[c]{@{}l@{}} the only young people who \\possibly will enjoy it  \end{tabular} & [neg] & \begin{tabular}[c]{@{}l@{}} earnhart 's \textit{film} is more about the \textcolor{red}{optimism} of \\a group of people who are struggling to give \\themselves a \textcolor{red}{better lot} in life than the ones \end{tabular} & [pos]\\
& 're down & [neg] &  \begin{tabular}[c]{@{}l@{}} ditched the \textcolor{blue}{saccharine} sentimentality \\of bicentennial man \end{tabular} & [pos]\\
& authentic to the core of his being & [pos] &  highlander & [pos]\\
\midrule
Representer & a guiltless \textit{film} & [pos] & seagal & [neg]\\
& \begin{tabular}[c]{@{}l@{}} it is different from others in its genre in that it is \\does not rely on dumb gags, anatomical humor, \\or \textit{character} cliches; it primarily relies \\on \textit{character} to tell its story \end{tabular} & [pos] & macaroni and cheese & [neg]\\
& between \textcolor{blue}{flaccid satire} and what & [pos] & grasping actors' workshop that it is & [neg]\\
\bottomrule 
\end{tabular}
}
\label{table: sst2-explanation-positive-full}
\end{table}

\begin{table}[H]
\caption{A wrongly predicted test example with negative sentiment in SST-2 interpreted by different instance attribution approaches.
}
\vskip 0.15in
\centering
\resizebox{1.0\linewidth}{!}{
\begin{tabular}{c|cc|cc}
\toprule
\multicolumn{4}{c}{Test input: \textcolor{red}{unflinchingly} \textcolor{blue}{bleak} and \textcolor{blue}{desperate}} & [neg]\\
\midrule
Attribution &\multicolumn{2}{|c|}{Most \textbf{helpful} training examples} & \multicolumn{2}{c}{Most \textbf{harmful} training examples}\\
\midrule
FreeShap & \textcolor{blue}{self-defeatingly} \textcolor{red}{decorous} & [neg] & excessively \textcolor{blue}{quirky} & [pos] \\
& \textcolor{blue}{oddly} \textcolor{red}{moving}  & [pos] &  it tight and \textcolor{blue}{nasty} & [pos]\\
& \textcolor{blue}{vicious} and \textcolor{blue}{absurd}  & [neg] &  \textcolor{blue}{damning} and damned \textcolor{red}{compelling} & [pos]\\
\midrule 
Influence & is completely at sea & [neg] & of businesses & [pos]\\ 
& more slowly & [neg]  & \begin{tabular}[c]{@{}l@{}} retread , with the emphasis on self-empowering \\schmaltz and big-wave surfing that gives pic its \\title an afterthought \end{tabular} & [neg]\\ 
& a guiltless film & [pos] & are more interesting ways of dealing with the subject & [neg]\\ 
 \midrule
TracIn & by surrounding us with hyper-artificiality & [neg] & \textcolor{red}{effective} film & [pos]\\
& \begin{tabular}[c]{@{}l@{}}standard thriller and drag audience \\enthusiasm to \textcolor{blue}{crush depth} \end{tabular}& [neg] &  are \textcolor{red}{crisp} and \textcolor{red}{purposeful} without overdoing it & [pos]\\
& i did n't care. & [neg] & lend some dignity to a \textcolor{blue}{dumb story} & [neg]\\
\midrule
Representer & a guiltless film & [pos] & seagal & [neg]\\
& their own mortality & [pos] & more slowly & [neg]\\
& \textcolor{blue}{peril} & [pos] & puzzling & [neg]\\
\bottomrule 
\end{tabular}
}
\label{table: sst2-explanation-negative-full}
\end{table}

\begin{table}[H]
\caption{The most helpful/harmful examples for SST-2 test datasets prediction interpreted by different attribution methods.}
\vskip 0.15in
\centering
\resizebox{1.0\linewidth}{!}{
\begin{tabular}{c|cc|cc}
\toprule
Attribution &\multicolumn{2}{|c|}{Most \textbf{helpful} training examples} & \multicolumn{2}{c}{Most \textbf{harmful} training examples}\\
\midrule
FreeShap & \begin{tabular}[c]{@{}l@{}}a dramatic comedy as pleasantly \textcolor{blue}{dishonest} \\and pat as any hollywood \textcolor{blue}{fluff}.  \end{tabular}  & [neg] & \begin{tabular}[c]{@{}l@{}} the paranoid claustrophobia of a submarine movie \\with the unsettling spookiness of the supernatural -- \\why did n't hollywood think of this sooner ? \end{tabular} & [pos] \\
& \begin{tabular}[c]{@{}l@{}} it 's anchored by \textcolor{red}{splendid} performances from \\an honored screen veteran and a \textcolor{red}{sparkling} \\newcomer who instantly transform themselves \\into a \textcolor{red}{believable} mother/daughter pair .\end{tabular}  & [pos] &  \begin{tabular}[c]{@{}l@{}}kids who are into this thornberry stuff \\will probably be in wedgie heaven \end{tabular}& [pos]\\
\midrule 
Influence & is about something very \textcolor{red}{interesting} and odd that & [pos] & it tight and \textcolor{blue}{nasty} & [pos]\\ 
& exercise in formula \textcolor{red}{crash-and-bash} action & [neg]  & \begin{tabular}[c]{@{}l@{}}its characterization of hitler and the \textcolor{blue}{contrived} \\nature of its \textcolor{blue}{provocative} conclusion \end{tabular} & [pos]\\ 
 \midrule
TracIn &  the only young people who possibly will enjoy it& [neg] & \begin{tabular}[c]{@{}l@{}} earnhart 's film is more about the \textcolor{red}{optimism} of a \\group of people who are  struggling to \\give themselves a \textcolor{red}{better} lot in life than the ones \end{tabular} & [pos]\\
& 're down & [neg] & \begin{tabular}[c]{@{}l@{}}ditched the \textcolor{blue}{saccharine} sentimentality of \\bicentennial man \end{tabular} & [pos]\\
\midrule
Representer & is about something very \textcolor{red}{interesting} and odd that& [pos] & \begin{tabular}[c]{@{}l@{}}4ever has the same \textcolor{red}{sledgehammer appeal} \\as pokemon videos , but ' \end{tabular}& [pos]\\
& is the one & [pos] & a guiltless film & [pos]\\
\bottomrule 
\end{tabular}
}
\label{table: sst2-explanation-global}
\end{table}

We further demonstrate the explanations for natural language inference tasks in Tab.~\ref{table: rte-explanation-positive-full}. [yes] means entailment, while [no] means not entailment. Each training example shown here is predicted correctly. The dataset used in this section is RTE, while the model used here is RoBERTa. 
Similar to the previous examples, the Shapley value yields consistently more insightful explanations since among the top three most significant training examples, there exists one that has almost the same semantics as the test example.

\begin{table}[H]
\caption{A correctly predicted example with an entailment pair in RTE interpreted by different instance attribution approaches.
}
\vskip 0.15in
\centering
\resizebox{1.0\linewidth}{!}{
\begin{tabular}{c|cc|cc}
\toprule
\multicolumn{4}{c}{\begin{tabular}[c]{@{}l@{}} Test input: \\
sentence1: As spacecraft commander for Apollo XI, the first manned lunar landing mission, \textcolor{blue}{Armstrong} was the \textcolor{blue}{first man to walk on the Moon}.\\ "That's one small step for a man, one giant leap for mankind." With these historic words, man's dream of the ages was fulfilled.\\
sentence2: \textcolor{blue}{Neil Armstrong} was the \textcolor{blue}{first man who landed on the Moon.} \end{tabular}} & [yes]\\
\midrule
Attribution &\multicolumn{2}{|c|}{Most \textbf{helpful} training examples} & \multicolumn{2}{c}{Most \textbf{harmful} training examples}\\

\midrule
FreeShap & \begin{tabular}[c]{@{}l@{}} sentence1: In 1867, Nobel obtained a patent on a special type of \\
nitroglycerine, which he called "dynamite". The invention quickly proved \\ its usefulness in building and construction in many countries.\\
sentence2: Alfred Nobel is the inventor of dynamite. \end{tabular} & [yes] & \begin{tabular}[c]{@{}l@{}} sentence1: Gabriel Garcia Marquez was a liberal thinker whose left-wing \\politics angered many conservative politicians and heads of state. \\His job as a reporter for the Cuban news agency Prensa Latina, in 1960, \\and and friendship with Fidel Castro resulted in his being ultimately \\ denied entry to the United States for political reasons\\
sentence2: Gabriel Garcia Marquez was a conservative politician. \end{tabular} & [no] \\
& \begin{tabular}[c]{@{}l@{}}sentence1: 
\textcolor{blue}{Neil Armstrong} was an aviator in the Navy and was chosen \\with the second group of astronauts in 1962.  Made seven flights in the \\X-15 program 1960 photo), reaching an altitude  of 207,500 feet.  Was \\backup command pilot for Gemini 5, command pilot for Gemini 8, backup \\ command pilotfor Gemini 11, backup commander for Apollo 8, and \\
commander for Apollo 11:  \textcolor{blue}{successfully completing the first moonwalk.} \\
sentence2: \textcolor{blue}{Neil Armstrong} was the \textcolor{blue}{first man who landed on the Moon.} \end{tabular}  & [yes] & \begin{tabular}[c]{@{}l@{}} sentence1: Evan Wolfson, the founder of the modern gay marriage \\ movement, tells the waiter he would like an iced decaf and "the usual."\\ Wolfson, one of  Time Magazine\'s Most Influential People in the World, is \\ a man who unflinchingly knows what he wants and stays his course,\\  whether it be in his choice of restaurant or in his choice of battle. \\And others always know when they see \\Evan coming what it is that he wants.\\
sentence2: Time Magazine was created by Evan Wolfson. \end{tabular} & [no]\\
& \begin{tabular}[c]{@{}l@{}}sentence1: Located just three miles from Tullamore and only 45 minutes \\ from the K Club, venue of the 2006 Ryder Cup, is Esker Hills, a genuine \\hidden gem and one of Irish golf's best kept secrets.\\
sentence2: The K Club will host the 2006 Ryder Cup. \end{tabular}  & [yes] &  \begin{tabular}[c]{@{}l@{}} sentence1: Hugh Jackman has done it all - almost. He's hosted the Oscars, \\danced on Broadway, flashed his winning grin and flexed his acting \\muscles, but Friday is the first time the Australian actor will claw his way \\ into movie theaters as the main attraction. In "X-Men Origins: Wolverine," \\ there's  no Halle Berry or  Nicole Kidman by his side to buoy the box office. \\This time it\'s just Jackman, his claws and a heap of Hollywood expectations \\ to the tune of \$150 million.\\
sentence2: Hugh Jackman will play the magician Harry Houdini. \end{tabular} & [no]\\

\midrule 
Influence & \begin{tabular}[c]{@{}l@{}} sentence1: Rather than deterring crime, capital punishment actually \\ increases the level of brutality in society.\\
sentence2: Capital punishment is a deterrent to crime.\end{tabular} & [no] & 
\begin{tabular}[c]{@{}l@{}} sentence1: A male rabbit is called a buck and a female rabbit \\ is called a doe, just like deer.\\
sentence2: A female rabbit is called a doe.\end{tabular}& [yes]\\ 
& \begin{tabular}[c]{@{}l@{}} sentence1: On Feb. 1, 1945, the Polish government made Warsaw \\ its capital, and an office for urban reconstruction was set up.\\
sentence2: Warsaw remained Poland's capital after the war.\end{tabular} & [no]  & 
 \begin{tabular}[c]{@{}l@{}} 
 sentence1: Bountiful arrived after war's end, sailing into San Francisco \\Bay 21 August 1945. Bountiful was then assigned as hospital ship \\
 at Yokosuka, Japan, departing San Francisco 1 November 1945.\\
sentence2: Bountiful reached San Francisco on 1 November 1945.\end{tabular}& [no]\\
& \begin{tabular}[c]{@{}l@{}} sentence1: Vera Beers, of Olivet, recently was named employee \\ of the month at Standard Printing in Marshall.\\
sentence2: Vera Beers works for Standard Printing.\end{tabular} & [yes]  & \begin{tabular}[c]{@{}l@{}} sentence1: 'In other words, with its 2 million inhabitants, \\Slovenia has only 5.5 thousand professional soldiers.\\
sentence2: Slovenia has 5.5 million inhabitants.\end{tabular}& [no]\\ 

 \midrule
TracIn & \begin{tabular}[c]{@{}l@{}} sentence1: Martin was taken to New Orleans Ochsner Foundation Hospital, \\where nurse Jinny Resor said he was treated for dehydration.\\
sentence2: Martin works for the Ochsner Foundation Hospital.  \end{tabular} & [no] & \begin{tabular}[c]{@{}l@{}} 
sentence1: Prime Minister Mahmoud Abbas has offered 'the hand of peace'\\
to Israel after his landslide victory in Sunday's presidential election.\\
sentence2: Mahmoud Abbas has claimed victory in the presidential elections. \end{tabular} & [yes]\\
& \begin{tabular}[c]{@{}l@{}} sentence1: In other words, with its 2 million inhabitants, \\Slovenia has only 5.5 thousand professional soldiers.\\
sentence2: Slovenia has 5.5 million inhabitants.\end{tabular}  & [no] &  \begin{tabular}[c]{@{}l@{}} sentence1: Canada is officially a bilingual country but, with nearly 60\% of \\
the population speaking English as their mother-tongue, and only 24\%\\ speaking French as their first language, some people are questioning \\ whether Canada is truly a bilingual nation or rather, a bilingual nation on \\ 
paper only. French is not the only linguistic minority in Canada, and some\\ 
of the languages spoken, in order of popularity, are Chinese, Italian,\\ German, Polish, Spanish, Portuguese, Punjabi, Ukrainian, Arabic, Dutch,\\ Tagalog, Greek, Vietnamese, Cree and Inuktitut.\\
sentence2: French is the most widely spoken language in Québec. \end{tabular} & [no]\\
& \begin{tabular}[c]{@{}l@{}} 
sentence1: With half of the vote counted, the ANC has 66\% of the \\
vote. Its nearest rival, the Democratic Alliance, has 16\%, \\
while ANC splinter-group the Congress of the People trails with 8\%.  \\
The results will see ANC leader Jacob Zuma elected as President of \\
South Africa when the National Assembly reconvenes in May.Provincial\\ elections are also being held, and the ANC looks likely to lose power \\
in the province of Western Cape to the Democratic Alliance. This will \\
be the first time an opposition party has won control of a provincial \\
parliament since the end of apartheid. The election campaign has \\
focused on crime, poverty, and the suitability of Zuma to be \\
President. Zuma was acquitted of rape in 2006, and corruption \\
charges against him were withdrawn shortly before the election after\\
prosecutors found the charges had been politically motivated.\\
sentence2: Jacob Zuma is the leader of the ANC party.\end{tabular}  & [yes] &  \begin{tabular}[c]{@{}l@{}} sentence1: Christopher Reeve, an actor and director who became an \\
inspiration worldwide after being paralyzed in a horse riding accident, died \\
Sunday of heart failure.\\ 
sentence2: Christopher Reeve had an accident. \end{tabular}& [yes]\\

\midrule
Representer & \begin{tabular}[c]{@{}l@{}} sentence1: A male rabbit is called a buck and \\ a female rabbit is called a doe, just like deer.\\
sentence2: A female rabbit is called a doe.\end{tabular} & [yes] & \begin{tabular}[c]{@{}l@{}} sentence1: In other words, with its 2 million inhabitants, Slovenia has \\ only 5.5 thousand professional soldiers.\\
sentence2: Slovenia has 5.5 million inhabitants. \end{tabular}  & [no]\\
& \begin{tabular}[c]{@{}l@{}} sentence1: A small bronze bust of Spencer Tracy sold for £174,000.\\
sentence2: A small bronze bust of Spencer Tracy made £174,000. \end{tabular} & [yes] &  \begin{tabular}[c]{@{}l@{}} 
sentence1: Bountiful arrived after war's end, sailing into San Francisco \\
Bay 21 August 1945. Bountiful was then assigned as hospital ship at\\ Yokosuka, Japan, departing San Francisco 1 November 1945.\\
sentence2: Bountiful reached San Francisco on 1 November 1945. \end{tabular} & [no]\\
& \begin{tabular}[c]{@{}l@{}} sentence1: In other words, with its 2 million inhabitants, \\Slovenia has only 5.5 thousand professional soldiers.\\
sentence2: Slovenia has 2 million inhabitants.\end{tabular} & [yes] & grasping actors' workshop that it is & [no]\\
\bottomrule 
\end{tabular}
}
\label{table: rte-explanation-positive-full}
\end{table}

\subsection{Extension to Data Selection}
\label{app: data-selection}
As elaborated in the main paper, we examine the potential of FreeShap in data selection. In this setting, we have a training set, a separate test set (validation set in the conventional sense, we refer to it as the test set to be consistent with the main paper) used for computing instance scores, and a held-out set (test set in the conventional sense) used for reporting the performance of the model trained on the selected data subset. Note that the setting of data selection is different from data removal as described in Sec.~\ref{sec: application_instance_attribution} in that data selection uses a separate held-out set to evaluate the performance of the selected data subset and hence is more applicable to selecting data in real-life applications. The setting of data removal is meant to validate the effectiveness of the instance attribution and hence is designed for analytical purposes. We conduct experiments on the MR dataset as its labeled held-out set is easily accessible. To evaluate FreeShap for data selection, we conduct two types of data selection tasks: data selection with addition and data selection with removal. The addition involves sequentially adding the data points with the highest score (included in the main paper), followed by evaluating the performance of the model on the held-out dataset retrained on the selected data; while the removal involves sequentially removing the data points with the lowest scores in order from the training set. 
Intuitively, in data selection with addition, a good data selection scheme should result in faster improvements in performance; while in data selection with removal, a good data selection scheme should result in improved performance or slower degradation of performance.
We supplement the results for data addition in LLama 2 in Tab.~\ref{table: mr_dataselection_bert}. FreeShap achieves faster improvements, validating its effectiveness as a data selection scheme in both models. However, for the removal setting in Tab.~\ref{table: mr_dataselection_llama}, FreeShap is less competitive. We hypothesize that FreeShap is computed primarily based on the marginal contributions of small subsets due to TMC sampling and hence is less effective in removing data points from large datasets w.r.t. the held-out set. 
Enhancing FreeShap for such scenarios remains a future research. 

\begin{table}[h]
\caption{
\textbf{Data Selection with Addition}:  Difference between the model performance of \textit{no data} (i.e., 0-shot) and the model performance when training on a \textit{subset that contains data points with high instance scores}. The subset size is given as a percentage of the full set. The higher the difference, the better the approach. 
}
\centering
\begin{tabular}{c|c|c|c|c|c}
\toprule
\multicolumn{6}{c}{BERT} \\
\midrule
Subset size &  2\% &  4\% &  6\% &  8\% &  10\% \\
FreeShap     & 0.1951 & \textbf{0.2111} & \textbf{0.2148} & \textbf{0.2167} & \textbf{0.2223}  \\ 
Influence   & 0.0272 & 0.0647 & 0.0393 & 0.0647 & 0.0750 \\
TracIn      & 0.1370 & 0.1904 & 0.2008 & 0.1548 & 0.1970 \\
Representer & 0.1182 & 0.1351 & 0.1388 & 0.0619 & 0.1238 \\
Random      & \textbf{0.1970} & 0.1951 & 0.2073 & 0.2017 & 0.1529 \\
\midrule
\multicolumn{6}{c}{Llama2} \\
\midrule
Subset size  &  2\% &  4\% &  6\% &  8\% &  10\%   \\
FreeShap     &  \textbf{0.0816} & \textbf{0.1398} & \textbf{0.1754} & \textbf{0.2186} & \textbf{0.2458}  \\ 
Influence   &  -0.0038 & -0.0038 & -0.0038 & -0.0038 & -0.0038\\ 
TracIn      &  0.0638 & 0.1144 & 0.1304 & 0.1773 & 0.2092  \\ 
Representer &  -0.0019 & 0.0000 & 0.0019 & 0.0066 & 0.0056   \\ 
Random      &  0.0638 & 0.1041 & 0.1529 & 0.1792 & 0.1951 \\ 
\bottomrule 
\end{tabular}

\label{table: mr_dataselection_bert}
\end{table}

\begin{table}[h]
\caption{\textbf{Data Selection with Deletion}: 
Difference between the model performance of \textit{all data} and the model performance when training on a \textit{subset that removes data points with low instance scores}. The subset size is given as a percentage of the full set. The smaller the difference, the better the approach. If the value is positive, which means the removal of low-value data improves the performance and is even better. 
}
\centering
% \vskip 0.15in
\begin{tabular}{c|c|c|c|c|c}
\toprule
\multicolumn{6}{c}{BERT} \\
\midrule
Subset size &  98\% &  96\% &  94\% &  92\% &  90\% \\
FreeShap     &  -0.0066 & \textbf{-0.0009} &  -0.0019 & -0.0150 & -0.0122 \\ 
Influence   &  -0.0038 & \textbf{ -0.0009} & -0.0019 &  -0.0056 &  -0.0141 \\
TracIn      &  \textbf{-0.0009} & -0.0056 &\textbf{-0.0009} & -0.0093 & -0.0113  \\
Representer & -0.0028  & -0.0066 & -0.0046 & -0.0066 & \textbf{-0.0066}  \\
Random      &  -0.0046 & -0.0094 & -0.0103 & \textbf{-0.0047} & -0.0188  \\

\midrule
\multicolumn{6}{c}{Llama2} \\
\midrule
Subset size &  98\% &  96\% &  94\% &  92\% &  90\%  \\
FreeShap     &  -0.0018 & -0.0009  & -0.0009 & -0.0018& 0.0009 \\ 
Influence   & \textbf{0.0028}  & \textbf{0.0019} & \textbf{0.0} & \textbf{0.0}  & \textbf{0.0019} \\ 
TracIn      & 0.0  & -0.0009  & -0.0009 & -0.0019 & -0.0009 \\ 
Representer &  -0.0028 & -0.0047 & -0.0084 & -0.0103 & -0.0122 \\ 
Random      &  -0.0056 & -0.0103 & \textbf{0.0} & -0.0009  & -0.0009\\ 
\bottomrule 
\end{tabular}

\label{table: mr_dataselection_llama}
\end{table}

\subsection{More Experiments for Multi-class on MNLI Dataset}
\begin{figure*}[!htb]
\centering
\includegraphics[width=12cm]{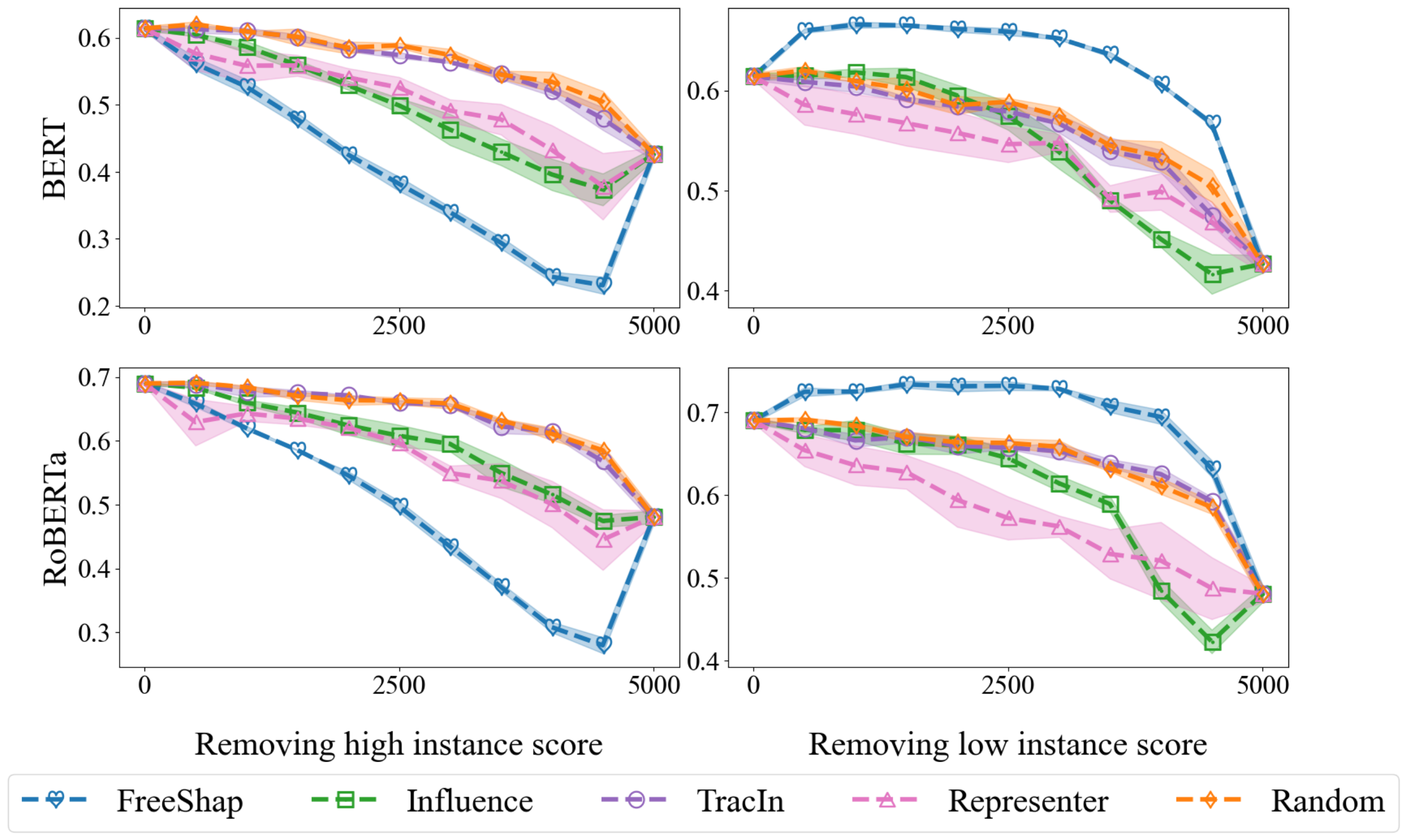}
\caption{\textbf{Data Removal}: The test accuracy on models retrained on subsets obtained by iteratively removing 10\% of the data, either from the highest or the lowest instance score. Faster degradation is preferable for high-score removals, while improvement or slower degradation is ideal for low-score removals. Overall, the scores from FreeShap are better correlated with test performance.
}
\label{fig: data_removal_mnli}
\end{figure*}

\begin{figure*}[!htb]
\centering
\includegraphics[width=8cm]{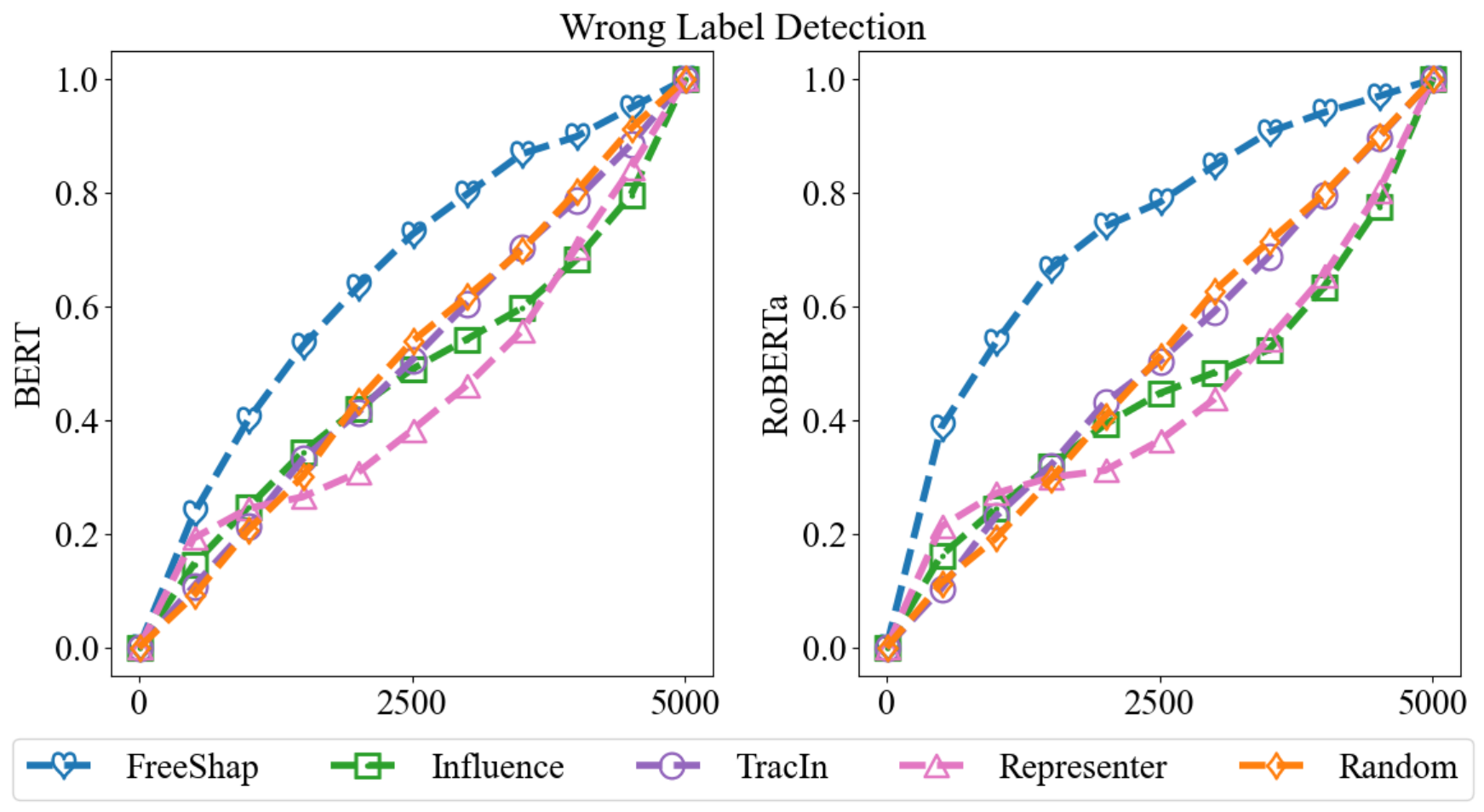}
\caption{\textbf{Wrong Label Detection}: It shows the detected percentage of poisoned data from the subsets with the lowest instance scores. For the MNLI dataset, FreeShap leads to the earliest identification of mislabeled instances.
}
\label{fig: data_poison_mnli}
\end{figure*}

For multi-class classification, we perform experiments on MNLI datasets (with $3$ classes). Regarding data removal, as shown in Fig.~\ref{fig: data_removal_mnli}, performance drops more rapidly when removing examples with the highest scores from FreeShap, and removing data based on the lowest scores from FreeShap even improves the performance. In wrong label detection, as shown in Fig.~\ref{fig: data_poison_mnli}, FreeShap is highly efficient at pinpointing mislabeled data.

\subsection{Scalability to Larger Datasets}
\begin{figure*}[!htb]
\centering
\includegraphics[width=16cm]{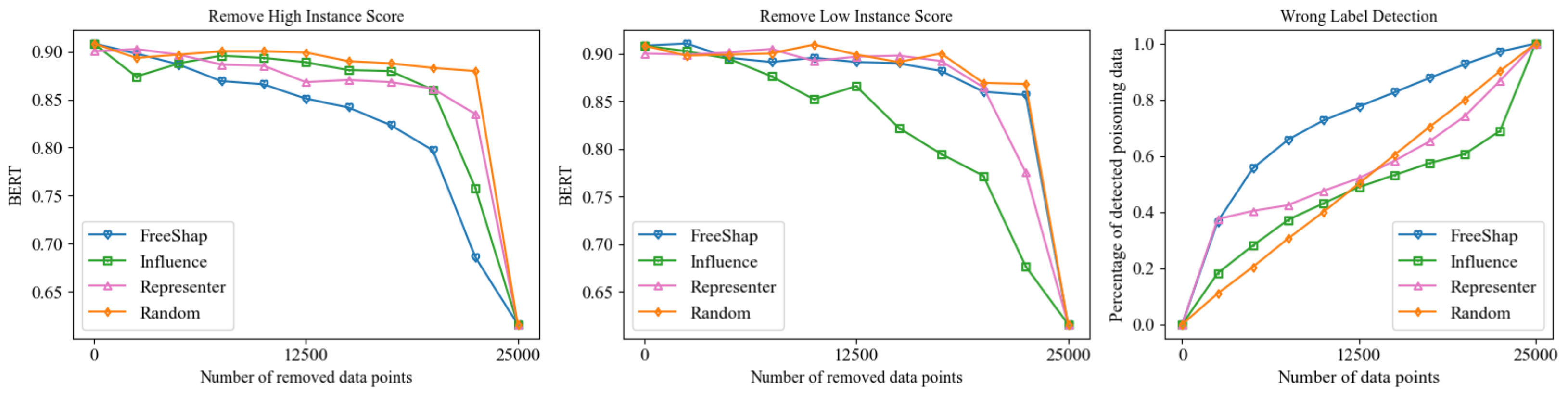}
\caption{\textbf{Data Removal and Wrong Label Detection}: The two figures on the left show the test accuracy on models retrained on subsets obtained by iteratively removing 10\% of the data. The rightmost figure shows the detected percentage of poisoned data from the subsets with the lowest instance scores. FreeShap is either superior or comparable in data removal while remaining effective at identifying wrong labels. 
}
\label{fig: sst2_25k}
\end{figure*}

To verify the scalability of FreeShap on larger datasets, we have extended our experiments on data removal and wrong label detection (described in Sec.~\ref{sec: experiment}) to 25k SST2 training data, demonstrating the practical advantages of FreeShap over other baselines at a larger scale. TracIn is not included due to the computational inefficiency and its weak performance in main paper experiments (Fig.~\ref{fig: data_removal}, ~\ref{fig: data_poison}). To further improve the efficiency of experiments, the hyperparameter of tolerance for FreeShap is further increased to $0.1$ and the Monte-Carlo iterations are increased to $800$. The influence function is computed with LiSSA and depth is $1000$. From Fig.~\ref{fig: sst2_25k}, when removing data with high instance scores, FreeShap outperforms other approaches. When removing data with low instance scores, the performance of FreeShap is comparable to other approaches. This is because the task is relatively simple as even random removal performs well. In addition, FreeShap performs well in identifying wrong labels.

\subsection{More Experiments for Llama2}
\label{app: llama_instance_exp}

We further empirically show FreeShap’s effectiveness as explanations, data removal, time efficiency, and a further upscaling to Llama2-13B here. 

\subsubsection{Test Prediction Explanations}
For the tables in this section: [pos] means positive sentiment class and predicted correctly, while [pos(neg)] means a positively labeled training example being wrongly predicted as negative, and vice versa. We use SST-2 in this section. 

We list more explanations using different instance attributions for the \textbf{most} helpful/harmful examples (with the highest/lowest scores) of a correct prediction in Tab.~\ref{table: sst2-explanation-positive-full-llama}. 
Like previous case studies, FreeShap consistently offers insightful explanations for the top three examples. 
Other methods may present drama-related examples but lack consistently decent performance.  

We provide explanations for an incorrect prediction in Table~\ref{table: sst2-explanation-negative-full-llama}, noting that the most supportive examples identified by FreeShap are syntactically similar to the test case, particularly in film-related content, and feature a mix of positive and negative words.

Additionally, we highlight training data with the highest/lowest instance scores relative to the test set in Table~\ref{table: sst2-explanation-global-llama}. Interestingly, most harmful training examples are incorrectly predicted. FreeShap and representer point both identify potentially wrongly labeled examples among the harmful examples, while the Influence function identifies potentially wrongly labeled examples among the helpful examples. Additionally, there's an overlap between FreeShap's helpful examples in the current table and those in Table~\ref{table: sst2-explanation-global}, as well as between the represented point's harmful examples across the two tables.

Last but not least, we also notice that the influence function and representer point tend to provide wrongly predicted training examples in explanations, aligning with ~\citet{pruthi2020estimating}'s experiments on the computer vision dataset. 

\begin{table}[H]
\caption{A correctly predicted example with positive sentiment in SST-2 interpreted by different attribution approaches on prompt-based fine-tuned \textbf{Llama2} model.}
\vskip 0.15in
\centering
\resizebox{1.0\linewidth}{!}{
\begin{tabular}{c|cc|cc}
\toprule
\multicolumn{4}{c}{Test input: ... a \textcolor{red}{magnificent} \textit{drama} \textcolor{red}{well worth} tracking down . } & [pos]\\
\midrule
Attribution &\multicolumn{2}{|c|}{Most \textbf{helpful} training examples} & \multicolumn{2}{c}{Most \textbf{harmful} training examples}\\
\midrule
FreeShap & a \textcolor{red}{consummate} \textit{actor} incapable of being boring & [pos] &\begin{tabular}[c]{@{}l@{}} others will find their humor-seeking dollars \\best spent elsewhere.\end{tabular} & [neg] \\
& \textcolor{red}{moving} and \textcolor{red}{solidly entertaining} & [pos] &  and \textcolor{blue}{undemanding} armchair tourists & [neg]\\
& \textcolor{red}{increasingly diverse}  french \textit{director} & [pos] &  \textcolor{blue}{bogs down} in a \textcolor{blue}{surfeit} of characters and & [neg]\\
\midrule 
Influence & most of it given to children & [pos(neg)] & \begin{tabular}[c]{@{}l@{}} 4ever has the same \textcolor{red}{sledgehammer} \\ \textcolor{red}{appeal} as pokemon \textit{videos} , but \end{tabular} & [pos(neg)]\\ 
& \begin{tabular}[c]{@{}l@{}} 's because the laughs come from fairly \\ basic \textit{comedic} constructs \end{tabular} & [pos(neg)]  & \begin{tabular}[c]{@{}l@{}}  kids who are into this thornberry stuff will \\probably be in wedgie heaven \end{tabular} & [pos(neg)]\\ 
& is consistently & [pos(neg)] & \begin{tabular}[c]{@{}l@{}}\textcolor{red}{definitely funny} stuff, \textcolor{blue}{but} it's more of \\the `laughing at' variety than the `laughing with.' \end{tabular}& [pos(neg)]\\
 \midrule
TracIn & \textcolor{red}{superb} & [pos] & the book 's \textcolor{blue}{irreverent} energy , and scotches most & [neg]\\
& what--and has all the dramatic weight of a raindrop & [neg] &  develop her own \textit{film} language & [pos]\\
& scored and powered by a set of \textcolor{red}{heartfelt} \textit{performances} & [pos] &  something of a public service -- & [pos]\\
\midrule
Representer & \begin{tabular}[c]{@{}l@{}} you 've endured a long workout without \\your pulse ever racing \end{tabular} & [pos(neg)] & \textcolor{blue}{self-deprecating, biting} and \textcolor{red}{witty} feature & [neg]\\
& an hour 's worth of actual material & [pos(neg)] & on its \textcolor{blue}{taut} \textit{performances} and \textcolor{blue}{creepy} atmosphere & [neg]\\
& \begin{tabular}[c]{@{}l@{}}beyond the \textcolor{red}{cleverness}, the \textcolor{blue}{weirdness} and the \\\textcolor{red}{pristine} camerawork, one hour photo is a sobering \\meditation on why we take pictures. \end{tabular}& [pos(neg)] & \begin{tabular}[c]{@{}l@{}} the ethos of the chelsea hotel may shape  \\ hawke 's  artistic aspirations , but  he has n't yet\\ coordinated his own dv poetry with the \\ beat he hears in his soul \end{tabular} & [neg]\\
\bottomrule 
\end{tabular}
}
\label{table: sst2-explanation-positive-full-llama}
\end{table}

\begin{table}[H]
\caption{A wrongly predicted example with positive sentiment in SST-2 interpreted by different attribution approaches on prompt-based fine-tuned \textbf{Llama2} model.}
\vskip 0.15in
\centering
\resizebox{1.0\linewidth}{!}{
\begin{tabular}{c|cc|cc}
\toprule
\multicolumn{4}{c}{Test input: like leon, it's \textcolor{blue}{frustrating} and still oddly  \textcolor{red}{likable}. } & [pos(neg)]\\
\midrule
Attribution &\multicolumn{2}{|c|}{Most \textbf{helpful} training examples} & \multicolumn{2}{c}{Most \textbf{harmful} training examples}\\
\midrule
FreeShap & \begin{tabular}[c]{@{}l@{}}it's an \textcolor{blue}{effort} to watch this movie, but it \textcolor{red}{eventually }\\\textcolor{red}{pays off} and is \textcolor{red}{effective} if you stick with it  \end{tabular} & [pos(neg)] & ended \textcolor{red}{so} \textcolor{blue}{damned} \textcolor{red}{soon} & [neg] \\
& \begin{tabular}[c]{@{}l@{}} this is lightweight filmmaking, to be sure, but it's \\ \textcolor{red}{pleasant} enough -- and oozing with \textcolor{red}{attractive} men. \end{tabular}& [pos] &  \begin{tabular}[c]{@{}l@{}} if even the filmmakers didn't know what kind of \\movie they were making  \end{tabular} & [neg]\\
&\begin{tabular}[c]{@{}l@{}}, it still manages to string \\together enough \textcolor{red}{charming} moments to work.\end{tabular} & [pos] & \begin{tabular}[c]{@{}l@{}} the story passes time until it's time for an \textcolor{blue}{absurd} \\finale of twisted metal, fireballs and revenge.\end{tabular} & [neg]\\
\midrule 
Influence & \begin{tabular}[c]{@{}l@{}} 4ever has the same \textcolor{red}{sledgehammer appeal} as \\pokemon videos , but \end{tabular} & [pos(neg)] & \textcolor{blue}{blood-curdling} family intensity & [pos]\\ 
& \begin{tabular}[c]{@{}l@{}}kids who are into this thornberry stuff\\ will probably be in wedgie heaven. \end{tabular} & [pos(neg)]  & \begin{tabular}[c]{@{}l@{}}to feel-good , follow-your-dream \\hollywood fantasies\end{tabular} & [pos(neg)]\\ 
& \begin{tabular}[c]{@{}l@{}} have \textcolor{blue}{strip-mined} the monty \\formula \textcolor{blue}{mercilessly} since 1997.\end{tabular} & [pos(neg)] & \begin{tabular}[c]{@{}l@{}}an hour 's worth of actual material  \end{tabular}& [pos(neg)]\\
 \midrule
TracIn & \textcolor{red}{superb} & [pos] & the book 's \textcolor{blue}{irreverent} energy , and scotches most & [neg]\\
& what -- and has all the dramatic weight of a raindrop & [neg] &  develop her own film language & [pos]\\
& scored and powered by a set of \textcolor{red}{heartfelt} performances & [pos] &  something of a public service -- & [pos]\\
\midrule
Representer & \begin{tabular}[c]{@{}l@{}} \textcolor{blue}{self-deprecating}, \textcolor{red}{biting and witty feature}  \end{tabular} & [neg] & \begin{tabular}[c]{@{}l@{}} feel the screenwriter at every moment  \end{tabular} & [pos(neg)]\\
& on its \textcolor{blue}{taut} performances and \textcolor{blue}{creepy} atmosphere  & [neg] & \begin{tabular}[c]{@{}l@{}} the \textcolor{blue}{sentimental cliches} mar an otherwise\\ \textcolor{red}{excellent} film .\end{tabular} & [pos(neg)]\\
& \begin{tabular}[c]{@{}l@{}}the ethos of the chelsea hotel may shape hawke's \\artistic aspirations, but he hasn't yet coordinated his\\ own dv poetry with the beat he hears in his soul \end{tabular}& [neg] & \begin{tabular}[c]{@{}l@{}}kids who are into this thornberry stuff will\\ probably be in wedgie heaven.\end{tabular}  & [pos(neg)]\\
\bottomrule 
\end{tabular}
}
\label{table: sst2-explanation-negative-full-llama}
\end{table}

\begin{table}[H]
\caption{The most helpful/harmful examples for SST-2 test datasets prediction interpreted by different attribution methods on prompt-based fine-tuned \textbf{Llama2} model.}
\vskip 0.15in
\centering
\resizebox{1.0\linewidth}{!}{
\begin{tabular}{c|cc|cc}
\toprule
Attribution &\multicolumn{2}{|c|}{Most \textbf{helpful} training examples} & \multicolumn{2}{c}{Most \textbf{harmful} training examples}\\
\midrule
FreeShap & \begin{tabular}[c]{@{}l@{}}a culture clash comedy only half as clever as it thinks it is  \end{tabular}  & [neg] & \begin{tabular}[c]{@{}l@{}} \textcolor{red}{lush and beautifully} photographed ( somebody \\suggested the stills might make a nice coffee table book )  \end{tabular} & [pos(neg)] \\
& \begin{tabular}[c]{@{}l@{}} it's anchored by \textcolor{red}{splendid} performances from \\an honored screen veteran and a \textcolor{red}{sparkling} \\newcomer who instantly transform themselves \\into a \textcolor{red}{believable} mother/daughter pair.\end{tabular}  & [pos] &  \begin{tabular}[c]{@{}l@{}}finding the characters in slackers or their \\ \textcolor{blue}{antics amusing, let alone funny}  \end{tabular}& [pos(neg)]\\
\midrule 
Influence & \begin{tabular}[c]{@{}l@{}}\textcolor{blue}{far too polite} to scale the lunatic heights of\\ joe dante 's similarly styled gremlins \end{tabular}& [pos(neg)] & is as \textcolor{red}{delightful} as it is derivative. & [pos(neg)]\\ 
& \begin{tabular}[c]{@{}l@{}}4ever has the same \textcolor{red}{sledgehammer} \\ \textcolor{red}{appeal} as pokemon videos , but\end{tabular} & [pos(neg]  & his rental car & [pos(neg)]\\ 
 \midrule
TracIn & \textcolor{red}{superb} & [pos] & the book 's \textcolor{blue}{irreverent} energy , and scotches most & [neg]\\
& what -- and has all the dramatic weight of a raindrop & [neg] &  develop her own film language & [pos]\\
\midrule
Representer & a \textcolor{red}{masterpeice}& [pos(neg)] & \begin{tabular}[c]{@{}l@{}}have \textcolor{blue}{strip-mined} the monty formula \\\textcolor{blue}{mercilessly} since 1997\end{tabular}& [pos(neg)]\\
& to the nonconformist in us all & [pos(neg)] & \begin{tabular}[c]{@{}l@{}}4ever has the same \textcolor{red}{sledgehammer appeal} \\as pokemon videos , but\end{tabular}  & [pos(neg)]\\
\bottomrule 
\end{tabular}
}
\label{table: sst2-explanation-global-llama}
\end{table}

\subsubsection{Remove Low Instance Score}
\begin{figure}[!htb]
\centering
\includegraphics[width=15cm]{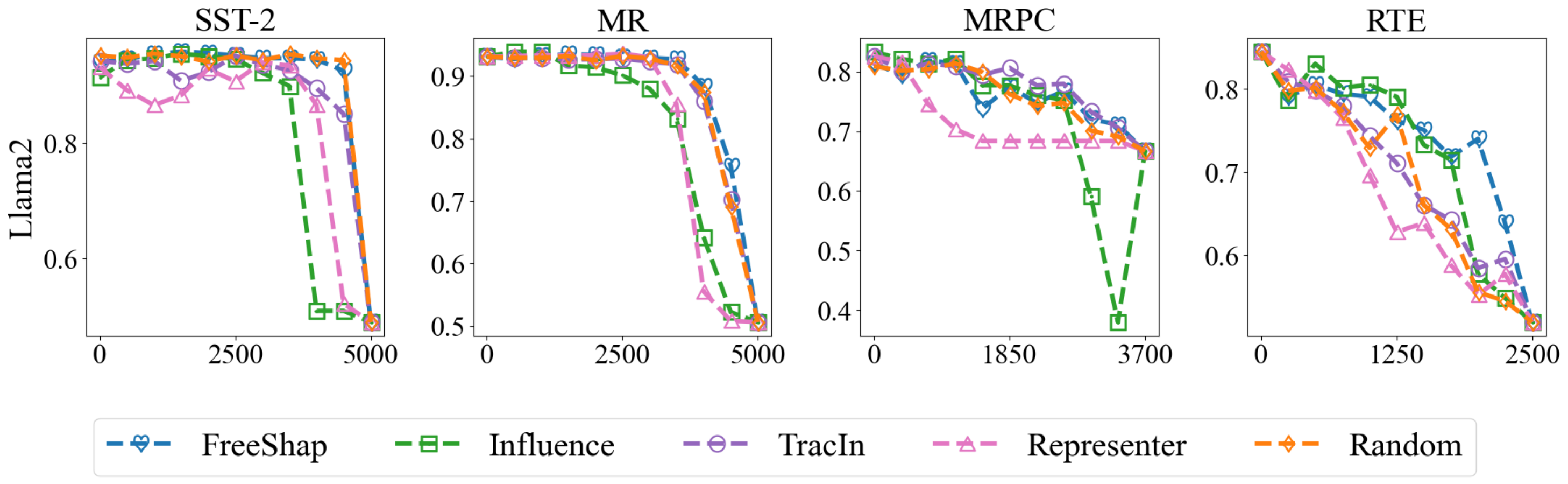}
\caption{
\textbf{Data Removal for Llama2}: When removing data from the lowest instance scores, the data points identified by FreeShap cause relatively slower degradation (slower is better). 
}
\label{fig: llama_data_removal}
\end{figure}

We include the results for removing low-instance sore points for Llama2. Fig.~\ref{fig: llama_data_removal} shows using FreeShap leads to a slower decrease in performance (most significant on the MRPC and RTE datasets) and even slight improvement for performance (in SST-2 and MR datasets).

\subsubsection{Time effiency}
\begin{figure}[H]
\centering
\includegraphics[width=8cm]{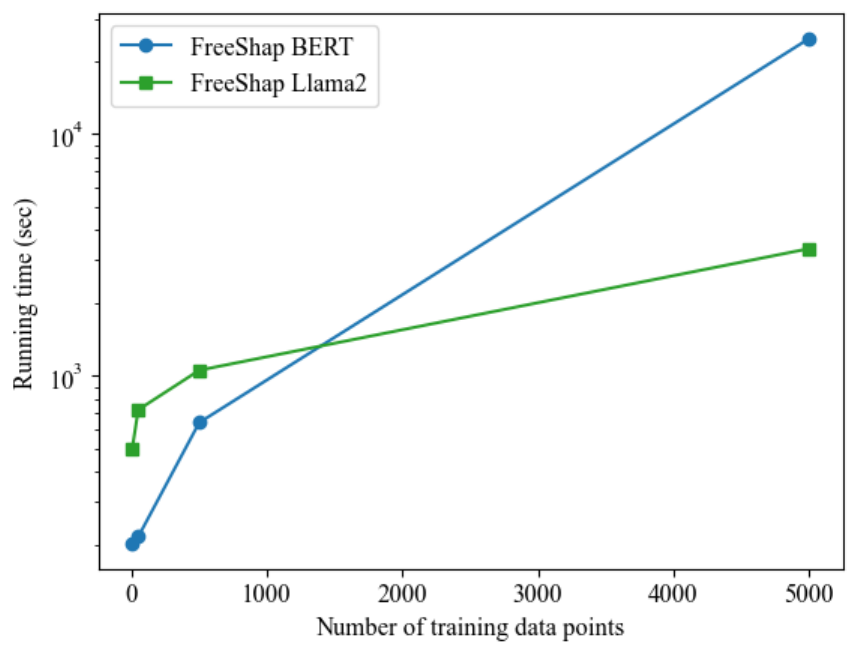}
\caption{Running time comparison for dataset SST-2 when the model is BERT and Llama2.}
\label{fig: time_llama}
\end{figure}

Fig.~\ref{fig: time_llama} shows the performance of BERT and Llama2-7B on the SST-2 dataset when using FreeShap with TMC and BI. The results indicate that FreeShap is still time efficient even with LLM. 
Initially, FreeShap takes a longer time to build the eNTK matrix for Llama2 due to the larger model size, but since the kernel regression avoids fine-tuning, the computational cost becomes independent of the model size. Interestingly, FreeShap on Llama2 can be even faster than BERT when the dataset becomes larger. This is because when TMC is applied, LoRA-finetuned Llama2 has better representation, which results in a more representative eNTK, and hence faster convergence (more pronounced diminishing marginal contribution) w.r.t. the number of training data and early termination for each Monte Carlo iteration when computing the Shapley value.

\subsubsection{Scaling up from Llama2-7B to Llama2-13B}
\begin{figure}[H]
\centering
\includegraphics[width=8cm]{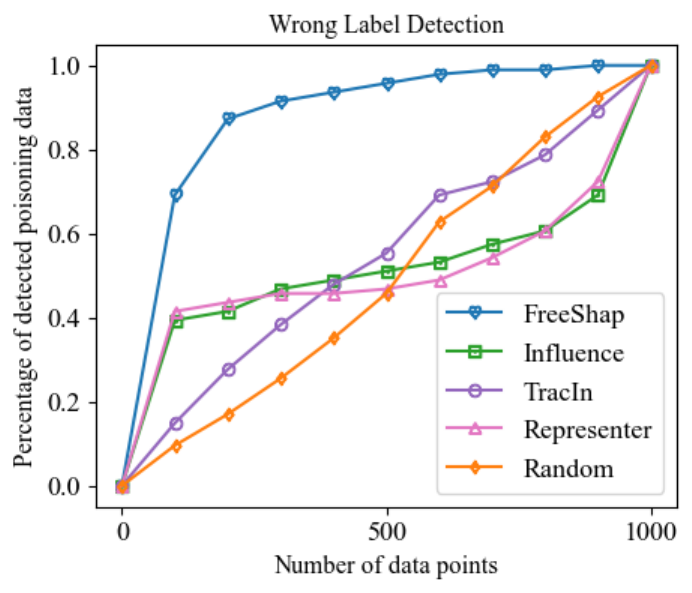}
\caption{Wrong Label Detection for dataset 1k SST-2 on Llama2-13b.}
\label{fig: llama13b}
\end{figure}

\begin{figure}[H]
\centering
\includegraphics[width=8cm]{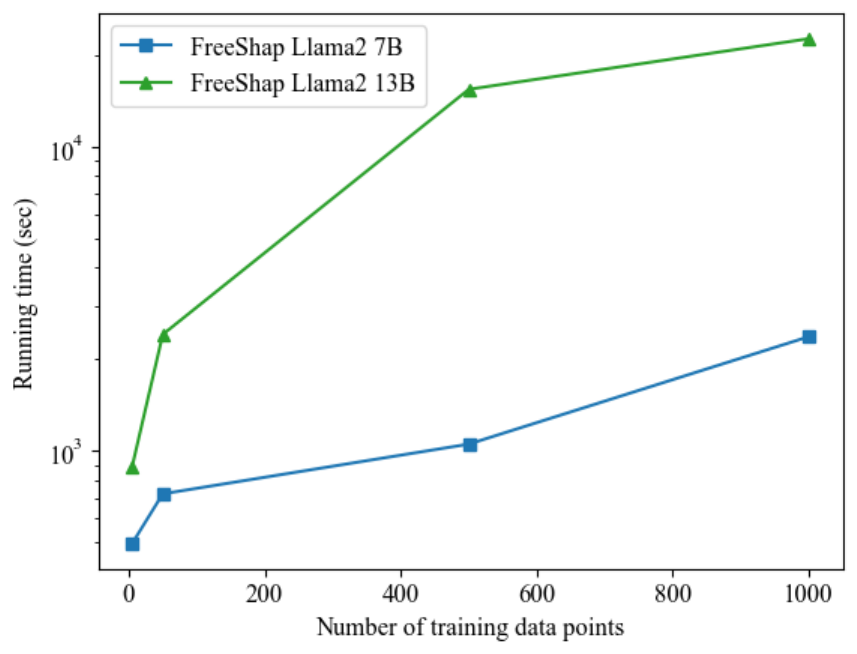}
\caption{Running time comparison for dataset SST-2 when the model is Llama2-7B and Llama2-13B.}
\label{fig: time_llama13b}
\end{figure}

To demonstrate the scalability of FreeShap w.r.t. the size of the model, we extend our experiments to the Llama2-13B model which is larger than the Llama2-7B model for wrong label detection on the SST2. We use a dataset with a size of 1k to demonstrate the scalability of the proposed method. As shown in Fig.~\ref{fig: llama13b}, FreeShap remains more effective in identifying mislabeled data with a larger model. We also demonstrate the time complexity when using Llama2-7B and Llama2-13B in Fig.~\ref{fig: time_llama13b}, showing the efficiency and scalability of FreeShap w.r.t. model size. Hence, our approach has the potential to be applied to larger models (70B or even more), and we encourage the community to explore this further with the right resources.

\section{Ablation Studies}

\subsection{Diminishing Influence of Data Point }
\begin{figure}[H]
\centering
\includegraphics[width=10cm]{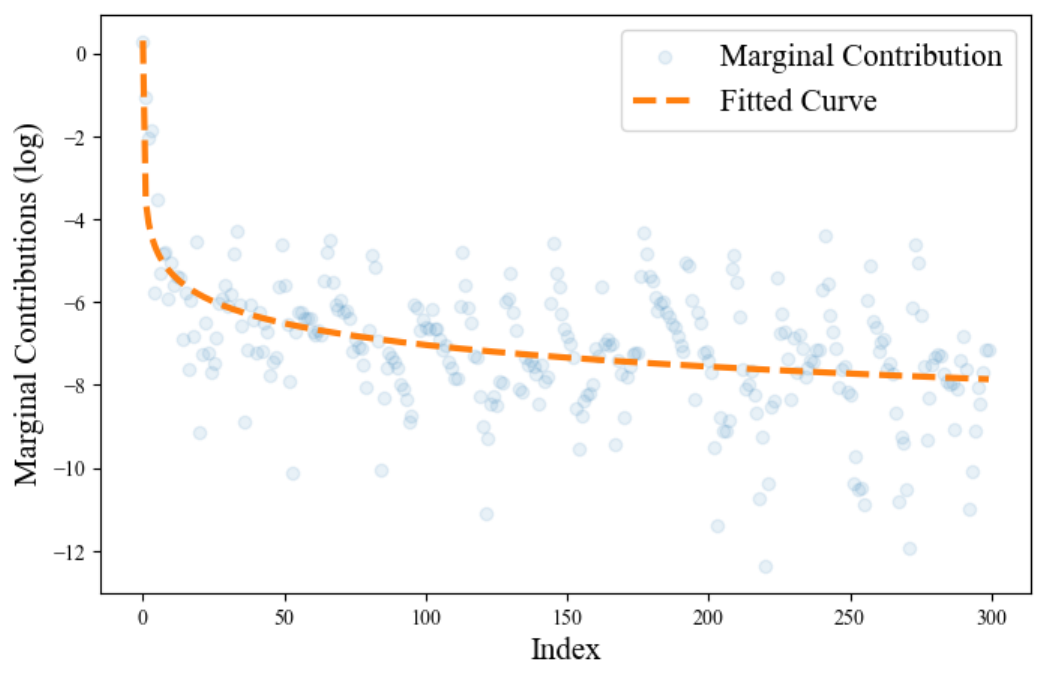}
\caption{\textbf{Diminishing Marginal Contribution}.}
\label{figure: diminishing_mc}
\end{figure}

We empirically examine the assumption of diminishing influence of a data point. Specifically, for a \textit{consistently harmful (or helpful) contributing} data point $z_i$, we examine if $|\tau_0| \geq \dots\geq |\tau_{n-1}|$. We randomly select a data point and compute the marginal contribution of this data point within subsets of varying sizes, ranging from 0 to 300. For each subset of size $k$, we calculate the mean marginal contribution from three sampled subsets. Figure~\ref{figure: diminishing_mc} plots the absolute mean marginal contribution of the data point across different subset sizes. The fitted curve illustrates a decreasing trend in the absolute marginal contribution, which is in accordance with our hypotheses and corroborated by previous studies~\citep{killamsetty2021glister, wang2021unified}.

\label{app: diminishing_mc}

\subsection{Expectation and Variance for Shapley value and LOO}
\label{app: mean_std_robust}
\begin{figure}[H]
\centering
    \subfloat[Mean of instance scores when the dataset is of size 500. The average of absolute mean among $50$ points is $0.50\text{e}{-3}$ for LOO and $1.20\text{e}{-3}$ for Shapley. ]{%
    	\includegraphics[width=7.5cm]{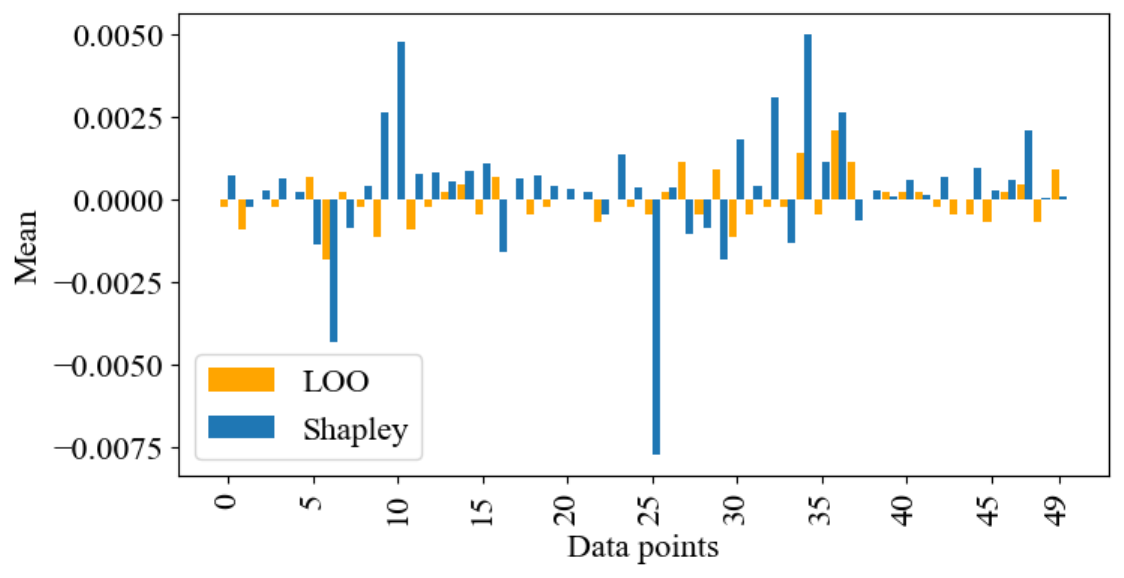}
	}\hspace{1em}
    \subfloat[Standard deviation of instance scores when the dataset is of size 500. The average of standard deviation among $50$ points is $1.89\text{e}{-3}$ for LOO and $0.61\text{e}{-3}$ for Shapley. ]{%
    	\includegraphics[width=7.5cm]{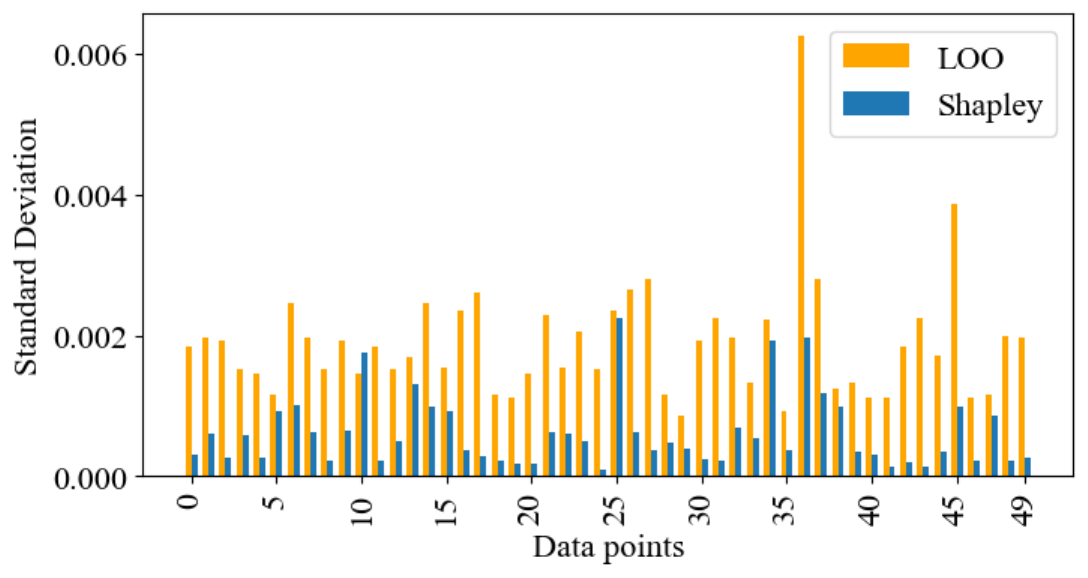}
	}\\

\caption{This figure shows each point's mean and standard deviation of five instance attribution scores computed by Shapley and LOO. The figure on the left demonstrates each point's mean of instance scores. Each index denotes a point's average instance score among five instance scores computed when the point is placed in five datasets. The figure on the right demonstrates each point's standard deviation of instance scores. Each index denotes a point's standard deviation of five instance scores. Matching with our analysis, the Shapley value has a higher expectation and a lower standard deviation.}
\label{fig: sst2_robust_mean_std}
\end{figure}

When comparing the robustness of Shapley and LOO, the upper bound is bounded by variables related to the expectation and variance. We can intuitively explain that the Shapley value is more robust as the Shapley value has a larger expectation and smaller variance, as illustrated in the remark~\ref{eqn: remark}. The relative relationship for expectation and variance between the Shapley value and LOO is further verified in the figure~\ref{fig: sst2_robust_mean_std}, where the figure on the left demonstrates that the Shapley value generally has a larger expectation, and the figure on the right demonstrates that the Shapley value has a smaller variance.

\subsection{Kernel Behavior of Prompt-based Fine-tuning}
\label{ablation: entk_kernel}
\begin{figure}[H]
\centering
\includegraphics[width=8cm]{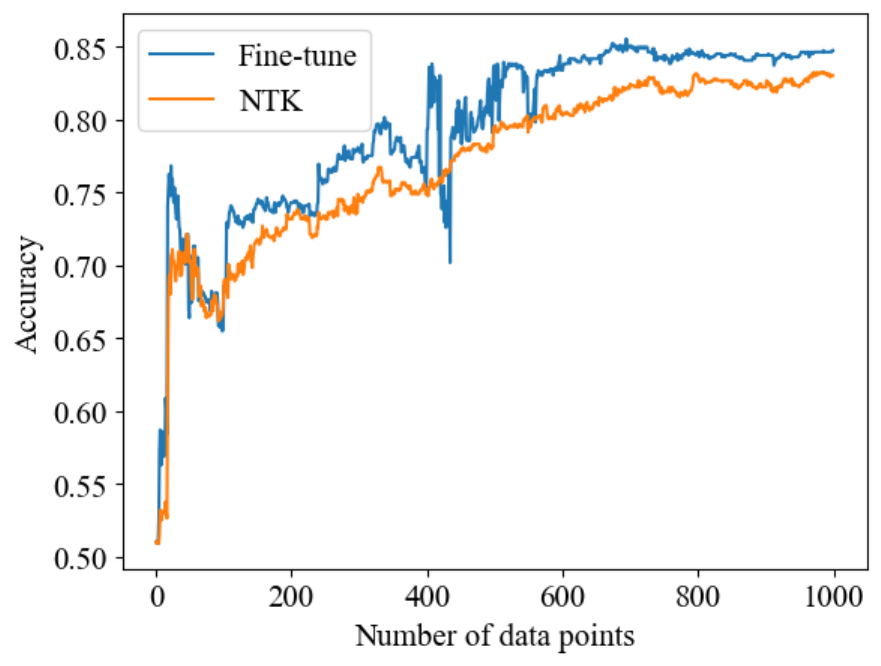}
\caption{We plot the test performance as the number of training points increases when using eNTK kernel regression and fine-tuning. The two lines have a Pearson correlation of $0.97$ and a Spearman correlation of $0.94$.}
\label{figure: entk_kernel}
\end{figure}

Previous work investigates the kernel behavior of prompt-based fine-tuning using experiments but on a few shot settings~\citep{pmlr-v202-malladi23a}. To further empirically verify the kernel behavior of prompt-based fine-tuning for each training data size, we conduct the following ablation study. We perform both prompt-based fine-tuning and eNTK kernel regression on training points with sizes ranging from 1 to 1k. We record test performance for each size and plot them in Figure~\ref{figure: entk_kernel}. At each training size, the eNTK (orange line) consistently provides a decent approximation to the fine-tuning method (blue line), evident from a small gap between the two curves. As the number of training points increases, both the eNTK and fine-tuning trends rise almost in tandem, indicating that eNTK offers a reliable approximation throughout the training process. 

\subsection{Ablation Studies for FreeShap and Influence on Different Approximations}
\label{app: ablation_appro}
\begin{figure}[H]
\centering
\includegraphics[width=13cm]{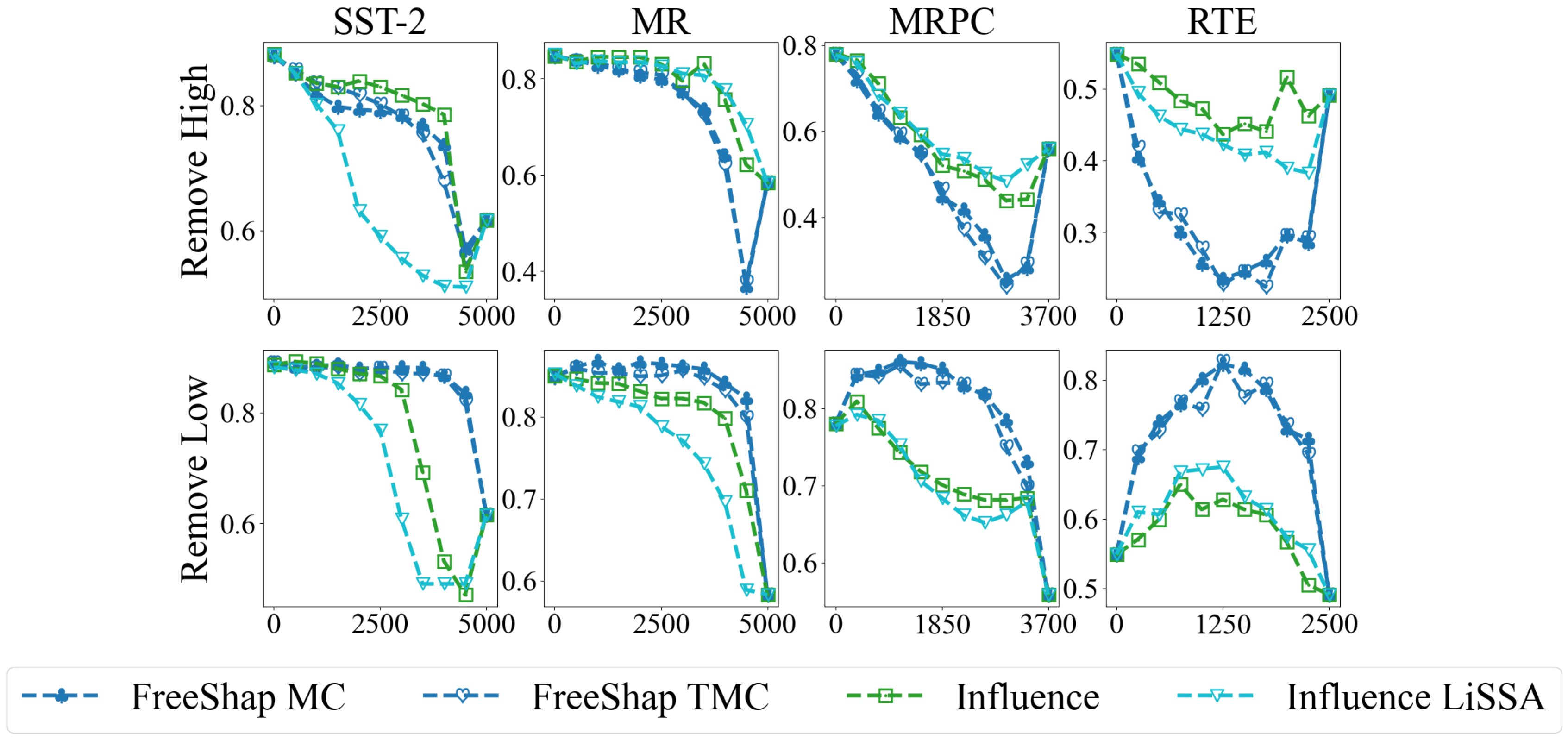}
\caption{\textbf{Data Removal}: Data removal performance comparison: FreeShap with TMC vs. FreeShap with MC show similar results. In contrast, the influence function with CG (used in the main paper) versus the influence function with LiSSA reveals disparities. The CG-approximated influence function excels at identifying low instance scores in most datasets but performs comparably or poorly in identifying high instance scores. }
\label{figure: appro_removal}
\end{figure}

\begin{figure}[H]
\centering
\includegraphics[width=13cm]{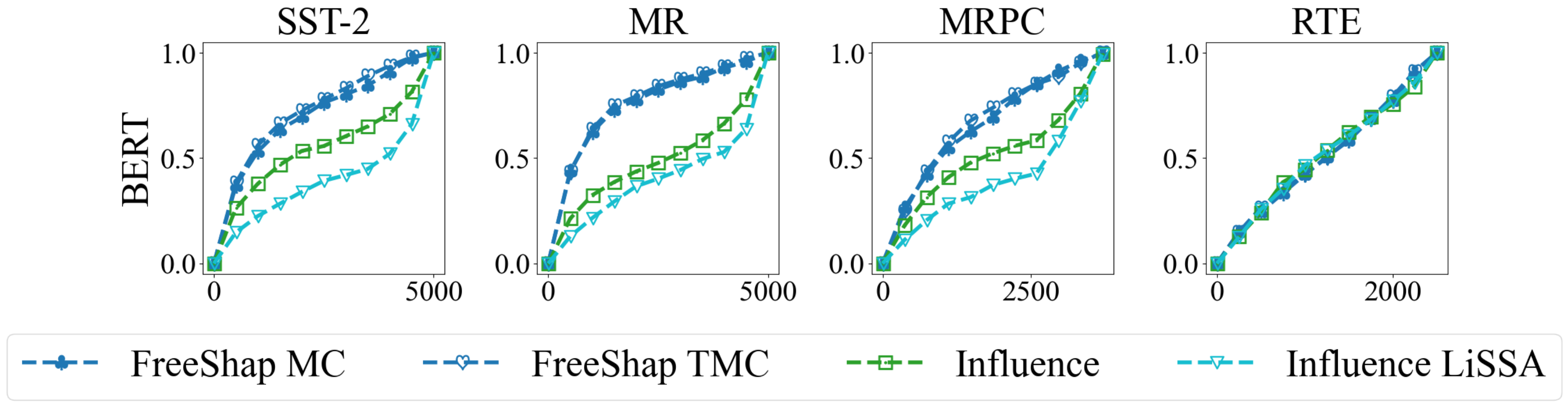}
\caption{\textbf{Wrong label detection}: Wrong label detection comparison: FreeShap with TMC vs. FreeShap with MC demonstrates similar efficacy in identifying mislabeled data. Further comparison between the influence function with CG (as used in the main paper) and the influence function with LiSSA shows the CG-approximated influence function to be more effective in pinpointing mislabeled examples.}
\label{figure: appro_poison}
\end{figure}

We compare the data removal performance (Fig.\ref{figure: appro_removal}) and wrong label detection (Fig.\ref{figure: appro_poison}) between FreeShap with TMC and FreeShap with MC, finding similar efficacy in identifying both high/low instance scores and mislabeled data.

Further, we evaluated the performance of the influence function with conjugate gradients (CG) (as used in the main paper) against the influence function with LiSSA in data removal  (Fig.\ref{figure: appro_removal}) and wrong label detection (Fig.\ref{figure: appro_poison}). The CG-based influence function demonstrated superior or comparable effectiveness in detecting low instance scores (slower degradation of performance is better) and mislabeled data, but it was comparable or less effective in identifying high instance scores (faster degradation of performance is better). Due to CG's proficiency in two of the three tasks, we choose CG for a more equitable comparison.

\section{Dataset Information}
\label{app: dataset}
Single sentence task involves classifying a single sentence independently to produce an output, like sentiment analysis (SST-2). The sentence pair task involves taking two sentences as input to determine a relationship between them, such as textual entailment or paraphrasing.

For the single-sentence task, we utilize two datasets: 
\begin{enumerate}
    \item The SST-2~\citep{socher-etal-2013-recursive} dataset is used for binary sentiment classification, aiming to categorize text snippets as either positive or negative. The dataset consists of movie reviews annotated with their corresponding sentiment labels. Dataset is retrieved from \url{https://huggingface.co/datasets/sst2}. 
    \item The Rotten Tomatoes movie review(MR) dataset is a collection of film reviews used for sentiment analysis tasks~\citep{pang-lee-2005-seeing}. The reviews are labeled as either positive or negative, based on the sentiment expressed by critics and audience members. The dataset is retrieved from \url{https://huggingface.co/datasets/rotten_tomatoes}. 
\end{enumerate}

For the sentence-pair task, we utilize three datasets: 
\begin{enumerate}
    \item The MRPC (Microsoft Research Paraphrase Corpus)~\citep{dolan-brockett-2005-automatically} dataset is a collection of sentence pairs labeled as either paraphrases of each other or not. It is commonly used for evaluating natural language processing tasks related to semantic textual similarity and paraphrase identification. Dataset is retrieved from \url{https://huggingface.co/datasets/glue/viewer/mrpc}. 
    \item The RTE (Recognizing Textual Entailment)~\citep{bentivogli2009fifth} dataset consists of sentence pairs labeled to indicate if the second sentence entails, contradicts, or is neutral concerning the first. It is widely used for natural language understanding tasks like textual entailment and inference. The dataset is retrieved from \url{https://huggingface.co/datasets/glue/viewer/rte}. 
    \item The MultiNLI (MNLI)~\citep{williams-etal-2018-broad} dataset is a large-scale, multi-genre natural language inference dataset designed for the training and evaluation of machine learning models on the NLI task. It contains pairs of sentences labeled as "entailment," "contradiction," or "neutral," drawn from a variety of sources, including fiction, government documents, and forums. The dataset is retrieved from \url{https://huggingface.co/datasets/glue/viewer/mnli}. 
    
\end{enumerate}

As our objective is to interpret the model predictions from the training data and obtain a better understanding of the dataset such as dataset debugging and mislabeled data detection, we directly evaluate all experiments on the test set. Since the test set label is not fully available (SST-2 and RTE), we use the validation set as the test set for simplicity. The necessary information for single sentence task datasets is in Tab.~\ref{table:single-sentence}, while the information for the sentence pair task datasets is in Tab.~\ref{table:sentence-pair}, and the test column in tables below is the information of validation datasets.

\begin{table}[h]
\centering
\caption{Detailed information about single sentence task dataset}
\begin{tabular}{c|cc|cc}
\toprule
\multirow{2}{*}{Dataset} & \multicolumn{2}{c|}{SST-2} & \multicolumn{2}{c}{MR}\\
 &  train & test & train & test \\
\midrule
length                   & 67.3k & 872   &  8.53k  &  1.07k   \\ 
min length of tokens     & 3     & 4     &  3      &   5      \\ 
max length of tokens     & 66    & 55    &  78     &   67     \\ 
average length of tokens & 13.32 & 25.16 &  27.37  &   27.66  \\ 
\bottomrule 
\end{tabular}
\label{table:single-sentence}
\end{table}

\begin{table}[h]
\centering
\caption{Detailed information about sentence pair task dataset}
\begin{tabular}{c|cc|cc|cc}
\toprule
\multirow{2}{*}{Dataset} & \multicolumn{2}{c|}{MRPC} & \multicolumn{2}{c|}{RTE} & \multicolumn{2}{c}{MNLI}\\
 &  train & test & train & test & train & test\\
\midrule
length                   & 3668 & 1725   &  2490  &  3000   &  392k &  9.8k\\ 
min length of premise     & 9     & 10     &  7      &   6      &   3  & 3\\ 
max length of premise     & 52    & 51    &  282     &   240     &   428  & 219 \\ 
average length of premise & 27.13 & 26.95 &  57.6  &   54.39  & 26.72 & 26.06\\ 
min length of hypothesis     & 10     & 10     &  5      &   4      &   3  & 3\\ 
max length of hypothesis     & 52    & 54   &  53     &   65     &   77  & 56 \\ 
average length of hypothesis & 27.12 & 27.03 &  13.59  &   13.24  & 14.19 & 14.12\\ 
\bottomrule 
\end{tabular}
\label{table:sentence-pair}
\end{table}

\end{document}